\pdfoutput=1
\documentclass{article}
\usepackage{PRIMEarxiv}
\usepackage[T1]{fontenc}
\usepackage[round]{natbib}
\setcitestyle{authoryear,round,citesep={;},aysep={,},yysep={;}}

\usepackage{amsmath,amsfonts,bm}

\def\eqref#1{equation~\ref{#1}}

\def\1{\bm{1}}

\DeclareMathAlphabet{\mathsfit}{\encodingdefault}{\sfdefault}{m}{sl}
\SetMathAlphabet{\mathsfit}{bold}{\encodingdefault}{\sfdefault}{bx}{n}

\usepackage{amsmath,amssymb,amsfonts}
\usepackage{booktabs}
\usepackage{array}
\usepackage{dcolumn}
\usepackage{tabularx}
\newcolumntype{L}[1]{>{\raggedright\arraybackslash}p{#1}}
\newcommand{\blocksep}{\specialrule{0.4pt}{0pt}{0pt}}%
\newcommand{\rowsep}{\specialrule{0.25pt}{0pt}{0pt}}%

\newcommand{\catrow}[2]{\multicolumn{#1}{@{}l@{}}{\colorbox{black!10}{\parbox{\dimexpr\linewidth-2\fboxsep\relax}{\rule{0pt}{2.1ex}\textbf{#2}}}}\\}
\newcommand{\apptab}{\footnotesize\renewcommand{\arraystretch}{1.02}\setlength{\tabcolsep}{3.2pt}}
\usepackage{multirow}
\usepackage[table]{xcolor}
\usepackage{soul}%
\usepackage{graphicx}
\usepackage{microtype}
\usepackage{pifont}
\usepackage{upquote}
\usepackage[most]{tcolorbox}
\usepackage{tikz}
\usetikzlibrary{arrows.meta,positioning,shapes.geometric,calc,fit,backgrounds}
\usepackage{enumitem}
\usepackage{xspace}
\usepackage{float}
\usepackage{placeins}%
\usepackage[normalem]{ulem}
\definecolor{ClaimRef}{HTML}{B71C1C}
\definecolor{ClaimRefLight}{HTML}{E8615F}
\definecolor{ClaimSup}{HTML}{1B5E20}
\definecolor{ClaimSupLight}{HTML}{4CAF50}
\newcounter{evidenceshade}
\newcommand{\Sclaim}[2]{{\ifodd#1\relax\color{ClaimSup}\else\color{ClaimSupLight}\fi\textbf{[S#1]}\,\uline{#2}}}
\newcommand{\Rclaim}[2]{{\ifodd#1\relax\color{ClaimRef}\else\color{ClaimRefLight}\fi\textbf{[R#1]}\,\uline{#2}}}

\soulregister\textdegree0
\makeatletter
\let\topicparagraph\paragraph
\renewcommand{\paragraph}{\@startsection{paragraph}{4}{\z@}{0.1pt}{-1em}{\normalsize\bf}}
\def\SOUL@hlpreamble{\setul{}{2.15ex}\let\SOUL@stcolor\SOUL@hlcolor\SOUL@stpreamble}
\makeatother
\newcommand{\bgmark}[2]{{\sethlcolor{#1}\hl{#2}}}
\newcommand{\SclaimBg}[2]{{\textcolor{ClaimSup}{\textbf{[S#1]}}\,\bgmark{ClaimBgSup}{#2}}}

\definecolor{ClaimBgSup}{HTML}{CDEFDD}
\definecolor{ClaimBgRef}{HTML}{F5B7B1}
\newcommand{\EsentR}[2]{{\textcolor{ClaimRef}{\textbf{[R#1]}}\,\bgmark{ClaimBgRef}{#2}}}
\newcommand{\RclaimBg}[2]{{\textcolor{ClaimRef}{\textbf{[R#1]}}\,\bgmark{ClaimBgRef}{#2}}}
\newcommand{\NclaimBg}[2]{{\textbf{[N#1]}\,#2}}
\newcommand{\EsentBg}[2]{{\textcolor{ClaimSup}{\textbf{[S#1]}}\,\bgmark{ClaimBgSup}{#2}}}
\newcommand{\Esent}[2]{\stepcounter{evidenceshade}{\ifodd\value{evidenceshade}\relax\color{ClaimSup}\else\color{ClaimSupLight}\fi\textbf{[#1]}\,\uline{#2}}}
\usepackage{afterpage}
\usepackage{needspace}
\usepackage{caption}

\newcommand{\fw}{Zephyrus-R1\xspace}%
\newcommand{\zeph}{\textsc{Zephyrus}\xspace}
\newcommand{\zephbench}{\textsc{ZephyrusBench}\xspace}

\newcommand{\wb}{WeatherBench~2\xspace}

\newcommand{\Direct}{Direct RL\xspace}%
\newcommand{\Curr}{Curriculum RL\xspace}%

\newcommand{\Q}{Q}%
\newcommand{\Rw}{R}%

\newcommand{\oss}{GPT-OSS\xspace}
\newcommand{\gemma}{Gemma\xspace}
\newcommand{\mistraljudge}{Mistral\xspace}
\newcommand{\qwenS}{Qwen3.5-4B\xspace}
\newcommand{\qwenL}{Qwen3.8-27B\xspace}

\definecolor{BoxBlue}{HTML}{2563EB}
\definecolor{BoxGreen}{HTML}{16A34A}
\definecolor{LinkBlue}{HTML}{1F3F8F}
\definecolor{BoxGold}{HTML}{CA8A04}
\definecolor{BoxPurple}{HTML}{9333EA}
\definecolor{BoxRed}{HTML}{DC2626}
\definecolor{BoxGray}{HTML}{6B7280}
\newtcolorbox{promptbox}[2][BoxBlue]{
  title=#2,
  colback=white,
  colframe=#1,
  colbacktitle=#1,
  coltitle=white,
  fonttitle=\bfseries,
  boxrule=1pt,
  arc=2mm,
  left=8pt, right=8pt, top=5pt, bottom=5pt,
  breakable,
  fontupper=\small\ttfamily,
  enhanced
}
\newtcolorbox{reportbox}[2][BoxGray]{
  title=#2,
  colback=white,
  colframe=#1,
  colbacktitle=#1,
  coltitle=white,
  fonttitle=\bfseries,
  boxrule=1pt,
  arc=2mm,
  left=8pt, right=8pt, top=8pt, bottom=8pt,
  breakable,
  fontupper=\small,
  enhanced
}

\newcommand{\rowsinput}[1]{\csname @@input\endcsname{#1.tex}}
\newcommand{\best}[1]{\textbf{#1}}

\definecolor{OurRow}{RGB}{230,239,252}
\definecolor{gain}{RGB}{0,122,61}
\definecolor{loss}{RGB}{200,80,0}
\newcolumntype{d}{r}
\makeatletter
\newcolumntype{B}[3]{>{\boldmath\DC@{#1}{#2}{#3}}c<{\DC@end}}
\makeatother
\newcommand{\BestNum}[1]{\textbf{#1}}
\newcommand{\BestAligned}[1]{\textbf{#1}}
\newcommand{\up}[1]{{\scriptsize\textcolor{gain}{+#1\%}}}

\usepackage{hyperref}
\usepackage{url}
\usepackage{wrapfig}
\hypersetup{colorlinks=true,linkcolor=LinkBlue,citecolor=LinkBlue,urlcolor=LinkBlue,
  pdftitle={Zephyrus-R1: Agentic Reinforcement Learning for Weather Reporting},
  pdfauthor={Fengfei Yu, Marshall Fisher, Isadora White, Sumanth Varambally, Yasaman Jafari, Shivam Yadav, Samuel M. Bartlett, Shawn Roj, Benjamin J. Moore, Michael D. Warner, Luyu Han, Zhirui Xia, Duncan Watson-Parris, Taylor Berg-Kirkpatrick, Yian Ma, Rose Yu},
  pdfsubject={Machine Learning; Weather Forecasting},
  pdfkeywords={Reinforcement learning, Agents, Large Language Models, Weather Science, Curriculum Learning}}

\title{Zephyrus-R1: Agentic Reinforcement Learning for Weather Reporting}

\author{
  Fengfei Yu, Marshall Fisher$^{*}$, Isadora White$^{*}$, Sumanth Varambally$^{*}$, Yasaman Jafari$^{*}$, Shivam Yadav, \\
  \bfseries Samuel M. Bartlett, Shawn Roj, Benjamin J. Moore, Michael D. Warner, Luyu Han, Zhirui Xia, \\
  \bfseries Duncan Watson-Parris, Taylor Berg-Kirkpatrick, Yian Ma, Rose Yu$^{\dagger}$ \\[4pt]
  University of California, San Diego \\
  \texttt{\{fey006, m1fisher, i2white, svarambally, yajafari, s4yadav, smbartlett, sroj,} \\
  \texttt{bem006, miwarner, l3han, z4xia, dwatsonparris, tberg, yianma, roseyu\}@ucsd.edu}
}

\begin{document}
\maketitle
{\renewcommand{\thefootnote}{}\footnotetext{$^{*}$Equal contribution.\quad $^{\dagger}$Correspondence: \texttt{roseyu@ucsd.edu}}}

\begin{abstract}

Reinforcement learning (RL) has emerged as a key method for training AI models in scientific applications, but RL remains difficult when outputs cannot easily be verified against a known ground-truth answer. Natural-language weather forecast reports are one such domain: they translate atmospheric data into assessments of weather evolution, hazards, and uncertainty for both the general public and key decision-makers. Unlike numerical weather-analysis questions, they admit many valid outputs, which limits deterministic verification. To address this, we develop \textbf{\fw}, a practical framework for RL on scientific weather forecast reporting. We focus on three key components of the agentic RL pipeline: reward design, curriculum learning, and an analysis of human-expert alignment. To encourage both report legibility and factual accuracy, we design a reward function that combines a meteorological rubric with bidirectional claim entailment, both evaluated against National Weather Service reference reports. Our framework optimizes this reward in a simulated tool-calling environment, enabling vision-language agents to gather and analyze forecast evidence across regions and lead times before producing a report. We use annotations from three meteorologists to audit the claim labels produced by our LLM judges. Finally, a curriculum helps small models first develop tool use and scientific reasoning through RL on verifiable weather-analysis questions, then optimize the open-ended report reward. After training Qwen3.5-4B with \textbf{\fw}, we improve the mean across six judge-metric scores by 26\% over the base model and achieve results competitive with frontier open-weight models. For Qwen3.8-27B, we improve the same mean by 13\% over the base model, by 6.7\% over Kimi-K3, and by 2.7\% over GPT-5.6-Luna.

\end{abstract}


\section{Introduction}
\label{sec:intro}

Reinforcement learning (RL) has driven remarkable progress in large language models (LLMs) for mathematics, programming, and tool-assisted reasoning \citep{shao2024deepseekmath,guo2025deepseekr1nature,jin2025searchr1,feng2025retool,zheng2025deepresearcher}. A key advantage of these domains is that success is often easy to verify: a proof is correct, a program passes its tests, or a numerical response is near its desired value. Many important scientific tasks, however, do not admit such straightforward evaluation. Even when model outputs can be grounded in scientific evidence, there may be no unique correct way to obtain a useful natural-language conclusion from it.  This raises a fundamental challenge: how do we use RL to optimize a scientific reasoning task where the objective itself is ambiguous?

Operational forecast discussions, or weather reports, provide an example of one such challenging domain. These reports convert numerical weather forecasts into actionable, legible reports designed for human use. They are relevant to a broad set of readers, including public decision-makers, business owners, and the general public. Producing a good report goes beyond making numerical predictions and demands synoptic (large-scale) weather pattern analysis, prioritization of reported phenomena, hazard identification, and communication of uncertainty. Different reports may produce different but equally useful summaries, while a fluent report can remain superficially plausible despite missing the most important event or mischaracterizing its cause. Due to the subjectivity involved, these tasks have traditionally been the domain of expert meteorologists. Thus, a core difficulty is not merely generating true meteorological text, but learning what constitutes a good operational report.

Recent work has explored the use of LLMs for meteorological reporting. \citet{lawson2025pixels} study multimodal interpretation of weather imagery, \citet{zheng2026weathersyn} instruction-tune a multimodal weather-report model, and \citet{higgins2026synopticbench} introduce a benchmark and domain-specific evaluation for synoptic reporting. Most closely related, \citet{singh2026afdbench} use RL to generate forecast discussions from structured WeatherNext~2 inputs \citep{Alet2026} with rewards for temperature accuracy, synoptic correctness, and format compliance. To our knowledge, our work is the first to tackle weather reporting as an open-ended, agentic reinforcement learning problem.

\begin{figure}
\centering
\includegraphics[width=\linewidth]{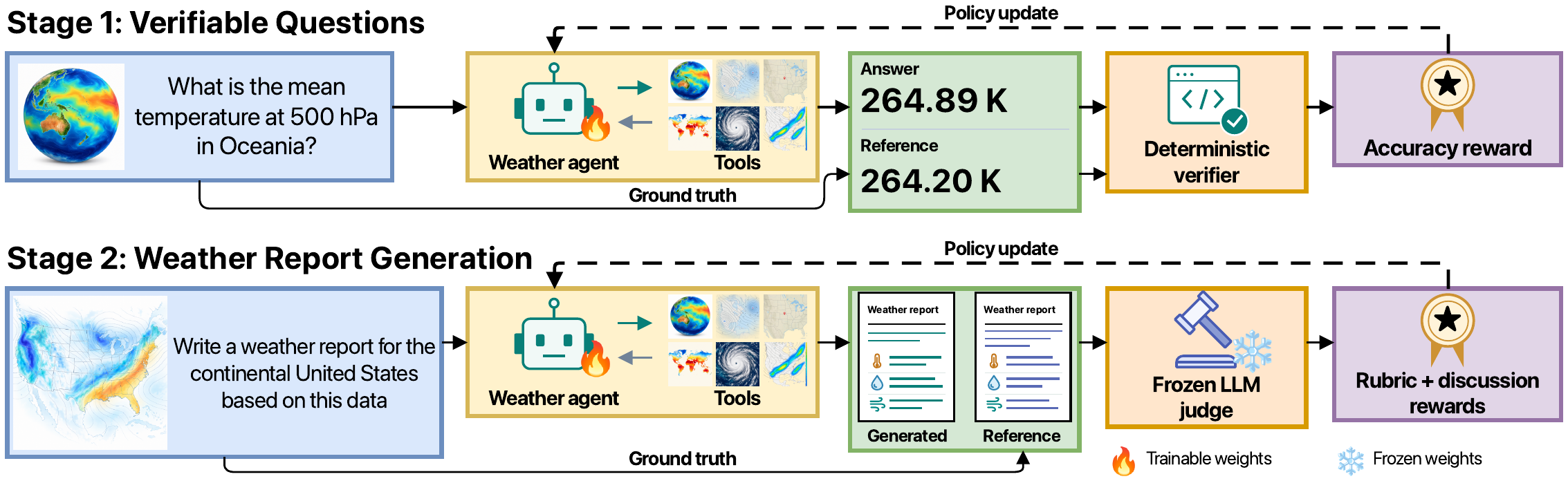}
\caption{\fw framework. Agents call weather tools, inspect returned data or errors, and produce answers or reports. Stage 1 rewards verified answers; Stage 2 scores reports on a frozen rubric and Discussion F1 via an LLM judge. Curriculum RL transfers the Stage-1 policy to report generation.}
\label{fig:overview}
\end{figure}

We develop \textbf{\fw} (Figure~\ref{fig:overview}), an agentic framework for weather-report generation in which a vision-language model acquires evidence through tools, synthesizes information across regions and lead times, and optimizes an open-ended report-level objective. The agent can inspect atmospheric data, forecasts, and plots before producing its report, which is scored using an LLM-evaluated six-criterion meteorological rubric and bidirectional claim entailment against National Weather Service reference reports. We audit the entailment judges' claim-label decisions against annotations from expert meteorologists. Because report generation simultaneously demands tool use, scientific reasoning, and open-ended synthesis, we train small models with a two-stage curriculum: RL first develops tool use and reasoning on weather-analysis questions with verifiable answers, then optimizes the report reward. This curriculum substantially improves \qwenS's tool-use capabilities (Section~\ref{sec:stage1-results}) and improves F1 scores over the base model by 67\% under the training LLM judge and by 24\% and 60\% under two independent judges. On \qwenL, \Curr is competitive with the strongest systems in our evaluation, outperforming Kimi-K3 on all metric-judge combinations and GPT-5.6-Luna on four of six combinations and in average overall score. Together, these results demonstrate that RL can improve open-ended scientific reporting when reward design, evidence acquisition, and training curriculum are designed jointly. Our contributions are as follows:

\begin{itemize}[leftmargin=1.2em,itemsep=1pt,topsep=2pt]
\item \textbf{Framework.} We introduce \fw, an RL framework for multi-turn, tool-using weather agents that combines weather-tool interaction, server-verifiable evidence acquisition, and judge-scored optimization of open-ended reports (Section~\ref{sec:framework}).
\item \textbf{Reward design.} We introduce a report-level reward that combines a meteorological rubric with bidirectional claim entailment to capture both report quality and factual coverage against a natural-language reference report (Section~\ref{sec:reward}).
\item \textbf{Domain expert evaluation.} Using annotations from three professional meteorologists, we compare expert and LLM claim labels. We find that weather-claim judgment is non-trivial and that LLM-expert agreement is comparable to inter-expert agreement (Section~\ref{sec:humans}).
\item \textbf{Curriculum learning.} We train tool use and analytical reasoning on verifiable weather-analysis questions, and study the effect of this curriculum-based approach compared to directly training on report generation (Section~\ref{sec:curriculum}).
\end{itemize}

\section{Related Work}
\label{sec:related}

\topicparagraph{Agents for science and weather.}
LLM agents that interleave reasoning with tool calls \citep{yao2023react,shinn2023reflexion,schick2023toolformer} have been applied across the sciences \citep{bran2024chemcrow,boiko2023coscientist,huang2025biomni}. In weather science, \zeph provides a Python environment over \wb with a neural forecaster as a tool and a 49-task benchmark \citep{varambally2026zephyrus}, while \textsc{SynopticBench} pairs forecast images with forecast discussions to evaluate vision-language models \citep{higgins2026synopticbench}.
Most closely related, \textsc{AFDBench} pairs structured WeatherNext~2 inputs with operational discussions and applies GRPO with domain-specific rewards for temperature accuracy, synoptic correctness, and format compliance \citep{singh2026afdbench}.
\textsc{ZephyrusBench} presents weather-science tasks as natural-language question-answer items, spanning data retrieval and computation, multi-step analysis, and forecast-report generation \citep{varambally2026zephyrus}. Stage 1 uses its verifiable weather-analysis questions, those with machine-checkable answers. Building on the \zeph environment, our policy acquires analysis and forecast fields through multi-turn tool use and transfers from these questions to report generation.

\topicparagraph{RL for tool use and rewards for open-ended generation.}
Group-relative policy optimization and related methods are widely used with verifiable rewards on tasks with checkable answers \citep{shao2024deepseekmath,yu2025dapo,guo2025deepseekr1nature,minimax2025m1}, and extensions to multi-turn tool use train the policy on its own execution traces with outcome rewards \citep{jin2025searchr1,feng2025retool,zheng2025deepresearcher,li2025websailor}.
However, reward hacking and tool-use collapse have been reported under weak supervision \citep{thaman2026rewardhacking,hao2026toolcollapse}.
For open-ended text, learned reward models and LLM judges \citep{ouyang2022instructgpt,zheng2023judging} are refined with rubrics \citep{gunjal2025rubrics,viswanathan2025checklists}, reference-derived chains \citep{jiang2026rlvrr}, mined rubrics \citep{wei2026qurl}, and hybrid verifier-plus-reward-model rewards \citep{tao2025hero}.
Our report stage uses a frozen judge with a fixed rubric and a claim-level metric, while the environment separately enforces the minimum tool trajectory needed for a weather report.

\topicparagraph{Curriculum learning.}
Ordering training from easy to hard \citep{bengio2009curriculum,narvekar2020curriculum} improves RL fine-tuning of LLM reasoning when difficulty is staged or adapted, with evidence on mathematics, planning, inductive reasoning, and other reasoning domains \citep{shi2025adarft,chen2025selfevolving,parashar2025e2h}.
These curricula span verifiable problem categories and domains. We study transfer from deterministically scored analysis to judge-scored reports, comparing \Curr with \Direct (Section~\ref{sec:ablation}).

\section{The \fw Framework}
\label{sec:framework}
\label{sec:setting}
\label{sec:method}

Our goal is to train a vision-language model-based agent to write accurate multi-day forecast discussions, given the state of the atmospheric weather conditions. To this end, we design an LLM-as-a-judge reward, audit the judges' claim labels against human experts, and optimize the agent using a two-stage curriculum learning framework.

\subsection{Setup}

\textbf{Problem setting.} We consider the task of generating multi-day operational forecast discussions from atmospheric weather data. Given access to the current atmospheric state and its evolution over the forecast horizon, the goal is to produce a discussion that summarizes the major weather patterns, their timing and location, and associated hazards. In this work, we focus on forecasts for the continental United States (CONUS).

To evaluate agents on this task and improve their performance, we follow the setup of \citet{varambally2026zephyrus}. We pair historical ERA5 reanalysis fields from WeatherBench~2 \citep{rasp2024weatherbench2,hersbach2020era5} with operational forecast discussions\footnote{\url{https://www.wpc.ncep.noaa.gov/discussions/hpcdiscussions.php?disc=pmdepd}} issued by the National Weather Service (NWS) on the corresponding dates. These human-written discussions serve as reference reports for training and evaluation.

\textbf{Agentic Environment.}
We formulate the task as a tool-using vision-language agent operating over atmospheric data. For each report-generation task, the agent receives the relevant date (excluding the exact year) and a collection of tools exposed through a Python API. It iteratively executes Python code to gather and analyze evidence from the numerical data to produce the forecast discussion. After each execution, the environment returns the resulting outputs and any execution errors, allowing the agent to refine its analysis or repair unsuccessful tool calls. When the agent creates plots, these are additionally returned as image observations, allowing it to reason directly over spatial weather patterns. The interaction continues until the agent produces its final forecast discussion.

\textbf{Toolset.}
Our environment builds on the toolset introduced in \textsc{Zephyrus} \citep{varambally2026zephyrus}. The agent can query WeatherBench~2 atmospheric fields, obtain geographic masks for regions of interest using the geolocator tool, and access historical climatology for computing anomalies. It also has access to a forecaster that returns atmospheric fields at requested forecast lead times. We use an ``oracle'' forecaster that returns the corresponding future ERA5 reanalysis field rather than the output of a numerical weather prediction model. This isolates the agent's ability to analyze atmospheric evolution and generate a forecast discussion from errors introduced by the underlying numerical forecast.

In this work, we augment the environment by adding vision capabilities to the agent. It can now leverage a plotting tool to plot arbitrary images during its code interactions. The execution environment then passes these figures back as image observations, so that the agent can inspect spatial patterns rather than infer them solely from printed arrays or statistics. We also add access to the TempestExtremes tool \citep{ullrich2017tempestextremes}, which allows the agent to detect and track large-scale weather systems such as tropical and extratropical cyclones.

\subsection{Reward design for the report stage}
\label{sec:reward}
\label{sec:motivation}

To evaluate the quality of the generated reports and train agents to improve on this task, we need a reward function $\Q(y, y^{*})$  that measures the quality of a generated report $y$ measured against a reference report $y^\ast$. Prior work on weather-report generation has found that standard text metrics such as BLEU and ROUGE are poor measures of forecast content quality, motivating more factual and domain-specific evaluation methods \citep{higgins2026synopticbench, varambally2026zephyrus}. We therefore seek to define a $\Q$ that rewards factual accuracy while penalizing hallucinations. To achieve this, we use two different LLM-as-a-judge metrics: a rubric score that compares $y$ against $y^{\ast}$ holistically, and a Discussion F1 score that measures claim-level factual precision and recall between $y$ and $y^{\ast}$.

\textbf{Rubric Score.} We first construct a meteorological rubric that captures the qualities of a strong forecast discussion. To do so, we generate faithful paraphrases and systematically corrupted variants of human-written reference reports, then use an LLM to iteratively refine criteria that separate the two groups. After review, we freeze a six-criterion rubric covering reference fidelity, spatiotemporal specificity, hazard prioritization, meteorological reasoning, thermal and precipitation-type plausibility, and calibrated use of detail. We then use this fixed rubric to score generated reports. The judge receives the generated report, the corresponding reference report, and the rubric, and assigns a score to each criterion. We average the six scores and normalize the result to $[0,1]$. The same rubric is used throughout training and evaluation. We provide the full rubric, scoring anchors, and judge prompt in Appendices~\ref{app:rubric} and~\ref{app:rubric-prompt}.

\textbf{Discussion F1.}
We also measure factual agreement at the level of individual claims. The LLM judge extracts atomic claims from both the generated and reference reports and labels each as SUPPORTED, REFUTED, or NEUTRAL with respect to the other report. We define
$\displaystyle P(y)=\frac{S_y}{S_y+F_y}, \; \mathrm{Rec}(y)=\frac{S_{y^{*}}}{N_{y^{*}}}$,  where $S_y$ and $F_y$ are the numbers of generated claims that are supported and refuted by the reference report, respectively, $S_{y^{*}}$ is the number of reference claims supported by the generated report, and $N_{y^{*}}$ is the total number of reference claims. Discussion F1 is the harmonic mean of precision and recall. This bidirectional claim-level metric follows the \zeph \citep{varambally2026zephyrus} ``discussion score.'' Unlike \zeph, which uses label probabilities derived from model logits, we use discrete LLM label decisions. The full scoring procedure is detailed in Appendix~\ref{app:entailment}.

\textbf{Report Grounding Requirement.} To ensure that the model actually examines the atmospheric data rather than hallucinating a plausible report, we require a minimum amount of successful tool use before assigning a judge-based reward. We maintain a binary grounding flag $g(\tau)$ for each trajectory $\tau$, which is set to one only if the agent has read a valid field from the current atmospheric state, obtained valid fields at a forecast lead, and received the resulting evidence in its context. Reports submitted before these conditions are satisfied are rejected, and the interaction continues. If $g(\tau)=0$ at the final turn, the rollout receives a fixed penalty, and the judge scores do not contribute to the training reward. The implementation is detailed in Appendix~\ref{app:agent-contract}.

\textbf{Judge-scored quality and terminal reward.} We combine the rubric and Discussion F1 scores as
\begin{equation}
\Q_t(y) = w_t\,\mathrm{rubric}(y) + (1-w_t)\,\mathrm{F1}(y),
\label{eq:quality}
\end{equation}
where $w_t$ controls the relative contribution of the two metrics. We update $w_t$ during training based on the variation of each score within the rollout group, so that neither metric dominates the reward. The final reward for training is
\begin{equation}
\Rw(\tau) =
\begin{cases}
-c, & \text{no usable report or } g(\tau)=0,\\
\Q_t(y) - p_{\text{fmt}}, & \text{otherwise},
\end{cases}
\label{eq:reward}
\end{equation}
where $c$ is a fixed penalty and $p_{\text{fmt}}$ applies only to a report recovered from an untagged forced-final reply. Since $\Q_t(y)\in[0,1]$ and $c > p_{\text{fmt}}$, any grounded report earns a higher reward than an ungrounded or missing one, so the policy cannot raise its return by skipping evidence acquisition. The weighting rule and constants are given in Appendix~\ref{app:reward-constants}.

\subsection{Human-expert audit of LLM claim labels}\label{sec:humans}
\label{sec:expert-audit}

\begin{wraptable}{r}{0.54\linewidth}
\vspace{-9pt}
\caption{Claim-label agreement on 100 claim-report pairs. Accuracy (Acc.) is exact label agreement; Within-1 excludes only SUPPORTED-REFUTED
reversals. For majority-vote comparisons, three-way ties are assigned NEUTRAL.}
\label{tab:expert-claim-agreement}
\centering
\footnotesize
\renewcommand{\arraystretch}{1.10}
\setlength{\tabcolsep}{1.8pt}
\begin{tabular*}{\linewidth}{@{\extracolsep{\fill}}lccc@{}}
\toprule
\textbf{Comparison} & \textbf{Acc.} & \textbf{$\kappa$} & \textbf{Within-1} \\
\midrule
GPT-OSS-Expert$_1$ & 0.370 & 0.097 & 0.920 \\
GPT-OSS-Expert$_2$ & 0.630 & 0.346 & \textbf{0.960} \\
GPT-OSS-Expert$_3$ & 0.570 & 0.226 & \textbf{0.960} \\
\addlinespace[2pt]
Expert$_1$-Expert$_2$ & 0.480 & 0.224 & 0.900 \\
Expert$_1$-Expert$_3$ & 0.460 & 0.214 & 0.860 \\
Expert$_2$-Expert$_3$ & 0.570 & 0.306 & 0.890 \\
\addlinespace[2pt]
GPT-OSS-Expert maj. & 0.630 & 0.327 & \textbf{0.960} \\
Gemma-Expert maj.   & 0.570 & 0.289 & 0.920 \\
Mistral-Expert maj. & 0.660 & \textbf{0.427} & 0.930 \\
LLM maj.-Expert maj.& \textbf{0.670} & 0.423 & 0.940 \\
\bottomrule
\end{tabular*}
\vspace{-7pt}
\end{wraptable}

To audit the claim labels produced by our LLM judges against the judgments of human domain experts, we asked three meteorologists to annotate the same 100 claims, using the SUPPORTED/REFUTED/NEUTRAL (S/R/N) definitions given to our LLM judges. To create the claim set, we sampled five report-generation tasks at random, drew report generators uniformly from \qwenS, \qwenL, and GPT-5.6-Luna, and gathered two groups for each task: generated claims scored against a reference report, and reference claims scored against a generated report. The experts were not informed which reports were reference or generated, nor which generating model produced each claim.

We also label the 100 claims with our three LLM judges (GPT-OSS-120B, Gemma-3-27B-IT, and Mistral-Small-3.2-24B) and compare the LLM labels with the expert labels, and the experts with one another. Table~\ref{tab:expert-claim-agreement} shows these results. We compute exact-match accuracy and Cohen's $\kappa$ over the categorical S/R/N labels, and Within-1 accuracy as the fraction of claims on which the two judgments are not opposite (i.e., not one SUPPORTED and one REFUTED). For majority-vote rows, we use the majority of three voters and assign three-way ties to NEUTRAL. Appendix~\ref{app:alignment} describes the audit protocol.

Our key finding is that this report claim annotation is a difficult task even for domain experts, with significant inter-expert disagreement. \oss agrees with the experts about as well as the experts agree with one another. In all LLM-expert comparisons, Within-1 accuracy exceeds 0.9, showing that almost no disagreements with the experts are full flips between SUPPORTED and REFUTED. Finally, we observe that accuracy is maximized by comparing three-judge LLM majority vote against human expert majority vote, suggesting that aggregating judgments helps reduce annotator-specific disagreement.

We use \oss as the RL training judge. On the same claim set, it is also the harshest judge, with a mean label score (SUPPORTED $=+1$, NEUTRAL $=0$, REFUTED $=-1$) of $0.05$, compared with $0.29$ for \mistraljudge and $0.57$ for \gemma. Its conservative labels provide a stringent training signal, while the independent judges test whether improvements transfer beyond the training judge.

\subsection{Curriculum Reinforcement Learning}
\label{sec:curriculum}

Direct report training must jointly learn tool use, weather analysis, and report composition from one scalar reward. We therefore use a two-stage curriculum: Stage 1 trains tool use on verifiable weather-analysis questions; Stage 2 optimizes report generation with the reward from Section~\ref{sec:reward}.

\subsubsection{Stage 1: Fine-tuning on verifiable weather-analysis questions}
\label{sec:stage1}

\paragraph{Accuracy reward.}
Stage 1 uses eight easy, machine-verifiable question templates from \zephbench~\citep{varambally2026zephyrus}. The questions ask the agent to compute regional statistics, identify spatial extrema and their timing, and detect climatological anomalies. Gold answers come from the benchmark metadata. Each question also specifies a required tool, which is the dataset for regional statistics and extrema and the climatology tool for anomaly questions. The verifier gives graded credit for numerical, location, time, and anomaly answers.
The reward is
\begin{equation}
r = s(\text{answer}, \text{gold}) \cdot \mathbf{1}[\text{required tool accessed}] - \lambda(\text{turns}),
\label{eq:stage1}
\end{equation}
where $s\in[0,1]$ is the graded verifier score and $\lambda$ applies a fixed deduction per turn beyond a turn allowance, up to a cap. An access counter verifies required-tool access. The question templates, score ladder, and turn penalty are detailed in Appendix~\ref{app:verifier}.

\Needspace*{17\baselineskip}
\begin{wrapfigure}{r}{0.5\linewidth}
\vspace{-8pt}
\centering
\includegraphics[width=\linewidth]{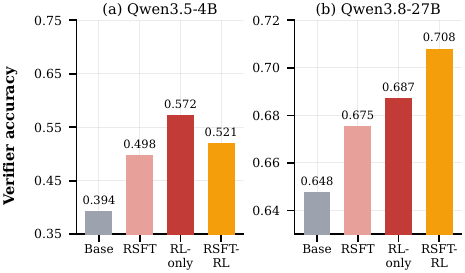}
\caption{Verifier accuracy on held-out validation weather-analysis questions for the base policy and the RSFT, RL-only, and RSFT-RL fine-tuned policies: (a) \qwenS; (b) \qwenL.}
\label{fig:stage1-curriculum}
\vspace{-7pt}
\end{wrapfigure}

\paragraph{Fine-tuning methods.}
Rejection-sampling fine-tuning (RSFT) retains full-credit, required-tool-access trajectories that pass infrastructure and output-format checks.%
\begin{figure}[!t]
\centering
\includegraphics[width=\linewidth]{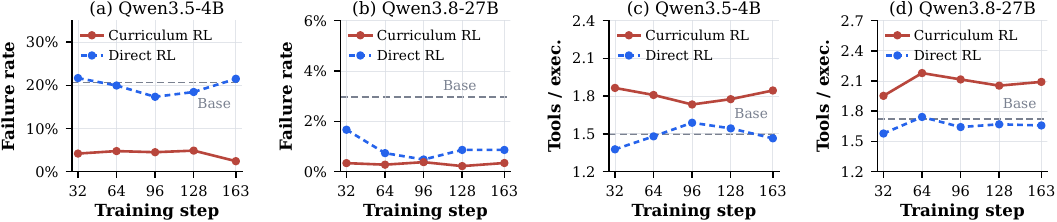}
\caption{Tool-use diagnostics on the report-stage test set during report RL. Failure rate is shown for 4B (a) and 27B (b), and Tools/exec. for 4B (c) and 27B (d). Appendix~\ref{app:selection} describes how both diagnostics are computed.}
\label{fig:effects}
\end{figure}%
\ Among eligible trajectories, RSFT keeps at most two per question, preferring fewer execution errors, turns, and tokens. Stage-1 RL optimizes Equation~\ref{eq:stage1}, starting from the RSFT checkpoint (RSFT-RL) or the base policy (RL-only).
All three methods raise verifier accuracy over the base policy at both scales (Figure~\ref{fig:stage1-curriculum}); the best checkpoints, RL-only for \qwenS (0.572) and RSFT-RL for \qwenL (0.708), initialize Stage 2. Because the base policies already access the required tool in over 99\% of rollouts, these gains reflect better use of and reasoning over the returned data. The questions also exercise component skills of report generation: quantifying field values, locating features in space, timing their evolution, and identifying anomalies relative to climatology. Stage 1 can therefore provide a stronger starting point and tool-use behavior for report RL. Appendix~\ref{app:curriculum} details the RSFT checks, schedule, selected checkpoints, results by answer type, and example trajectories.

\subsubsection{Stage 2: Reinforcement learning on the report generation task}
\label{sec:stage2}
Stage 2 trains on the report task with the reward of Equation~\ref{eq:reward}. Beyond these component skills, it requires forecast-tool use, hazard prioritization, and report composition. \Curr initializes the policy from the selected Stage-1 checkpoint and resets the optimizer. On the other hand, \Direct starts from the base policy and uses the same report data, prompt, reward rule, judge definitions, sampler, and report-stage length. The two routes therefore differ only in their initialization and the additional Stage-1 training used by \Curr.

\paragraph{Stage 1 reduces execution failures and broadens tool use.}
\label{sec:stage1-results}
Every Stage-2 report must first meet the grounding requirement of working analysis and forecast calls. This overlaps with Stage 1 in requiring data access, while forecast acquisition is specific to the report stage. We therefore compare the two routes on two diagnostics: the failure rate (the fraction of execution requests in a rollout that return no valid tool result) and Tools/exec. (the number of distinct tool families returning valid results per execution request). At every evaluated checkpoint and at both scales, \Curr has the lower failure rate and the higher Tools/exec. (Figure~\ref{fig:effects}). At step 163, the 4B failure rate is 2.4\% for \Curr versus 21.5\% for \Direct; at 27B, both routes are reliable (0.3\% versus 0.9\%), while Tools/exec. is 2.09 versus 1.66. Curriculum RL thus acquires evidence more reliably and broadly.

\subsection{Optimization}
\label{sec:optim}
Both stages use group-relative policy optimization \citep{shao2024deepseekmath} with the clipped importance-sampling policy optimization (CISPO) objective \citep{minimax2025m1}. Each rollout is a single sequence containing the full interaction history. For each prompt, we sample $K$ rollouts and compute group-centered advantages $A_k=r_k-K^{-1}\sum_j r_j$ without standard-deviation normalization, where $r_k$ is the stage reward of rollout $k$ (Equation~\ref{eq:stage1} or~\ref{eq:reward}). With $\mathcal{I}_k$ indexing the policy-generated tokens of rollout $k$, the sequence-mean CISPO loss is
\begin{equation}
\mathcal{L}(\theta)=-\frac{1}{K}\sum_{k=1}^{K} A_k\,\frac{1}{|\mathcal{I}_k|}\sum_{i\in\mathcal{I}_k}
\operatorname{sg}\!\big[\operatorname{clip}(\rho_{k,i},1-\epsilon_{\mathrm{low}},1+\epsilon_{\mathrm{high}})\big]\,
\log\pi_\theta(a_{k,i}\mid s_{k,i}),
\label{eq:cispo}
\end{equation}
where $\rho_{k,i}$ is the importance ratio between the current and sampling policies and $\operatorname{sg}$ stops gradients through the clipped ratio. Environment tokens are excluded from $\mathcal{I}_k$. Failure handling and hardware are described in Appendix~\ref{app:infra}.

\begin{table}[!t]
\caption{Weather-report results on the test events for the base policies and two \fw training routes. Every score averages three complete scoring passes; higher is better for both metrics, and bold marks the within-scale best. Green percentages give the relative change from the same-scale Base, computed from unrounded scores.}
\label{tab:main}
\centering
\small
\renewcommand{\arraystretch}{1.04}
\setlength{\tabcolsep}{1.0pt}
\resizebox{\linewidth}{!}{%
\begin{tabular}{@{}ll
  d@{\hspace{2pt}}l d@{\hspace{2pt}}l @{\hspace{2pt}}
  d@{\hspace{2pt}}l d@{\hspace{2pt}}l @{\hspace{2pt}}
  d@{\hspace{2pt}}l d@{\hspace{2pt}}l @{\hspace{2pt}}d@{\hspace{2pt}}l@{}}
\toprule
& & \multicolumn{4}{c}{\shortstack{\textbf{GPT-OSS}\\[-1pt]\textbf{120B}}} &
\multicolumn{4}{c}{\shortstack{\textbf{Gemma-3}\\[-1pt]\textbf{27B-IT}}} &
\multicolumn{4}{c}{\shortstack{\textbf{Mistral-Small}\\[-1pt]\textbf{3.2-24B}}} & & \\
\cmidrule(lr){3-6}\cmidrule(lr){7-10}\cmidrule(lr){11-14}
\textbf{Model} & \textbf{Route} &
\multicolumn{2}{c}{\textbf{Rub.}} & \multicolumn{2}{c}{\textbf{F1}} &
\multicolumn{2}{c}{\textbf{Rub.}} & \multicolumn{2}{c}{\textbf{F1}} &
\multicolumn{2}{c}{\textbf{Rub.}} & \multicolumn{2}{c}{\textbf{F1}} & \multicolumn{2}{c}{\textbf{Avg}} \\
\midrule
Qwen3.5-4B & Base & 0.082 & & 0.138 & & 0.507 & & 0.621 & & 0.391 & & 0.342 & & 0.347 & \\
 & \Direct & 0.123 & \up{51} & 0.217 & \up{57} & \BestNum{0.575} & \up{13}
 & 0.693 & \up{12} & \BestNum{0.453} & \up{16} & 0.477 & \up{40} & 0.423 & \up{22} \\
 & \Curr & \BestNum{0.129} & \up{57} & \BestNum{0.231} & \up{67} & 0.514 & \up{1}
 & \BestNum{0.768} & \up{24} & 0.441 & \up{13} & \BestNum{0.548} & \up{60} & \BestNum{0.438} & \up{26} \\
\midrule
Qwen3.8-27B & Base & 0.136 & & 0.224 & & 0.735 & & 0.715 & & 0.690 & & 0.537 & & 0.506 & \\
 & \Direct & 0.167 & \up{23} & \BestNum{0.318} & \up{42} & 0.742 & \up{1}
 & \BestNum{0.770} & \up{8} & 0.719 & \up{4} & \BestNum{0.631} & \up{18} & 0.558 & \up{10} \\
 & \Curr & \BestNum{0.260} & \up{91} & 0.312 & \up{39} & \BestNum{0.778} & \up{6}
 & 0.736 & \up{3} & \BestNum{0.728} & \up{6} & 0.615 & \up{15} & \BestNum{0.571} & \up{13} \\
\bottomrule
\end{tabular}}
\end{table}

\section{Experimental Results}

\subsection{Setup}
\label{sec:setup}

\paragraph{Data.}
Our reference reports are forecast discussions for the continental United States issued by the National Weather Service, the source of the \textsc{ZephyrusBench} report task~\citep{varambally2026zephyrus}. Written by operational forecasters, these discussions provide expert-quality targets without manual annotation. For report RL, we train on 2,608 events from 1995--2018 and evaluate on a test set of 197 temporally disjoint events from 2019--2020.
Stage 1 uses eight easy, machine-verifiable question templates from \textsc{ZephyrusBench}, covering regional statistics, geographic and temporal extrema, and one-sided, standardized, and percentile-envelope climatological anomaly detection.
Stage-1 RL draws 704 questions from a 710-question pool, with 125 disjoint questions held out for validation. Field coverage, report preprocessing, and report split-integrity checks are detailed in Appendix~\ref{app:data}.

\paragraph{Metrics and judges.}
\label{sec:metrics}
We report the rubric score and Discussion F1 (Section~\ref{sec:reward}), both in $[0,1]$; the rubric score rescales the mean of the six 1--10 criterion scores. GPT-OSS-120B is the training judge, while Gemma-3-27B-IT and Mistral-Small-3.2-24B are independent judges that apply the same rubric and fixed claim prompts to extract and label their own claims \citep{agarwal2025gptoss,gemmateam2025gemma3,mistral2025small32}. All judge calls use temperature zero, and each score in Tables~\ref{tab:main} and~\ref{tab:leaderboard} averages three scoring passes over the same frozen outputs. Appendix~\ref{app:judges} explains the repeated passes and compares the judges, and Appendix~\ref{app:entailment-prompts} gives the claim prompts.

\paragraph{Systems.}
For both \qwenS \citep{yang2025qwen3,qwenteam2026qwen35} and \qwenL \citep{qwenteam2026qwen38}, we compare the base model with \Direct and \Curr. Table~\ref{tab:leaderboard} also includes GPT-5.6-Luna \citep{openai2026gpt56} and eight open-weight models: Kimi-K3 \citep{moonshot2026kimik3}, DeepSeek-V4-Flash-Vision-Exp \citep{deepseek2026vision}, GLM-5.3-Flash \citep{zhipu2026glm5}, MiniMax-M3 \citep{minimax2026msa}, Qwen3.6-35B-A3B \citep{qwenteam2026qwen36}, Qwen3.5-122B-A10B and Qwen3.5-9B \citep{qwenteam2026qwen35}, and Gemma-4-31B-IT \citep{gemmateam2026gemma4}. Each model produces one report per test event.

\paragraph{Training setup.}
Both RL stages use CISPO with group-centered advantages, 16 prompts and four rollouts per step, temperature 0.6, zero KL regularization, and rank-64 LoRA on the language layers with the vision tower frozen \citep{hu2022lora}. Episodes allow at most 13 turns, the last forcing a final answer. Stage-1 RL runs for 44 steps with checkpoints selected by verifier credit on the validation set; the curriculum routes start report RL from 4B RL-only step 44 and 27B RSFT-RL step 24. Report RL runs one 163-step epoch with the reward of Equation~\ref{eq:reward}, re-estimating $w_t$ at every step and bounding it to $[0.2,0.8]$, with $c=0.05$ and $p_{\text{fmt}}=0.02$. Report-stage learning rates are $10^{-5}$ for 4B and $3\!\times\!10^{-5}$ for 27B. We report the terminal step-163 policy, so no test result selects a checkpoint or hyperparameter. We provide the remaining settings in Appendices~\ref{app:curriculum}--\ref{app:details}.

\subsection{Main results}
\label{sec:main-results}
\label{sec:ablation}

\paragraph{\fw improves report quality under all three judges.}
In Table~\ref{tab:main}, both routes improve every score at both scales. At 4B, \Direct and \Curr raise the six-score average by 22\% and 26\%, but they improve different aspects of the report: \Direct obtains slightly higher rubric scores under \gemma and \mistraljudge, whereas \Curr achieves higher F1 under all three judges, with the largest margins under \gemma (0.768 versus 0.693) and \mistraljudge (0.548 versus 0.477). This F1 advantage coincides with the much lower failure rate of the 4B curriculum agent (Section~\ref{sec:stage1-results}). At 27B, the averages rise by 10\% and 13\%, and the pattern reverses: \Curr achieves higher rubric scores under every judge, while \Direct retains slightly higher F1. The 27B curriculum agent also uses tools more broadly, yet its advantage appears in the rubric score rather than F1, so execution diagnostics alone do not determine which metric benefits. Overall, report RL alone accounts for most of the improvement at both scales, showing that the judge-based reward is an effective training signal for open-ended reports, and the curriculum adds a further 3--4 percentage points of relative gain (26\% versus 22\% at 4B; 13\% versus 10\% at 27B). The relative gains are largest under \oss, the training judge, but they persist under both independent judges, so they are not confined to the evaluator being optimized. Appendix~\ref{app:checkpoints} tracks both routes across training.

\begin{table}[t]
\caption{Leaderboard on the test events, ordered by Avg (higher is better). Every score averages three complete scoring passes. Shaded rows are Zephyrus-R1 models trained with Curriculum RL. Avg is the mean of the six unrounded scores.}
\label{tab:leaderboard}
\centering
\small
\renewcommand{\arraystretch}{1.04}
\setlength{\tabcolsep}{4pt}
\begin{tabular}{@{}r@{\hspace{6pt}}l d d<{\hspace{6pt}} >{\hspace{6pt}}d d<{\hspace{6pt}} >{\hspace{6pt}}d d<{\hspace{6pt}} >{\hspace{6pt}}d@{}}
\toprule
& & \multicolumn{2}{c}{\shortstack{\textbf{GPT-OSS}\\[-1pt]\textbf{120B}}} &
\multicolumn{2}{c}{\shortstack{\textbf{Gemma-3}\\[-1pt]\textbf{27B-IT}}} &
\multicolumn{2}{c}{\shortstack{\textbf{Mistral-Small}\\[-1pt]\textbf{3.2-24B}}} & \\
\cmidrule(lr){3-4}\cmidrule(lr){5-6}\cmidrule(lr){7-8}
\textbf{\#} & \textbf{Model} &
\multicolumn{1}{c}{\textbf{Rub.}} & \multicolumn{1}{c}{\textbf{F1}} &
\multicolumn{1}{c}{\textbf{Rub.}} & \multicolumn{1}{c}{\textbf{F1}} &
\multicolumn{1}{c}{\textbf{Rub.}} & \multicolumn{1}{c}{\textbf{F1}} &
\multicolumn{1}{c}{\textbf{Avg}} \\
\midrule
\rowcolor{OurRow}
1 & Qwen3.8-27B + \fw & \BestAligned{0.260} & \BestAligned{0.312} & \BestAligned{0.778} & 0.736 & \BestAligned{0.728} & 0.615 & \BestAligned{0.571} \\
2 & GPT-5.6-Luna & 0.170 & 0.297 & 0.713 & \BestAligned{0.801} & 0.698 & \BestAligned{0.660} & 0.556 \\
3 & Kimi-K3 & 0.168 & 0.286 & 0.736 & 0.732 & 0.721 & 0.570 & 0.536 \\
4 & DeepSeek-V4-Flash-Vision-Exp & 0.153 & 0.262 & 0.706 & 0.729 & 0.689 & 0.570 & 0.518 \\
5 & GLM-5.3-Flash & 0.156 & 0.224 & 0.728 & 0.726 & 0.699 & 0.567 & 0.517 \\
6 & MiniMax-M3 & 0.139 & 0.241 & 0.709 & 0.716 & 0.682 & 0.572 & 0.510 \\
7 & Qwen3.6-35B-A3B & 0.126 & 0.197 & 0.619 & 0.708 & 0.530 & 0.481 & 0.444 \\
8 & Qwen3.5-122B-A10B & 0.107 & 0.199 & 0.584 & 0.736 & 0.525 & 0.500 & 0.442 \\
\rowcolor{OurRow}
9 & Qwen3.5-4B + \fw & 0.129 & 0.231 & 0.514 & 0.768 & 0.441 & 0.548 & 0.438 \\
10 & Gemma-4-31B-IT & 0.136 & 0.190 & 0.558 & 0.702 & 0.490 & 0.465 & 0.424 \\
11 & Qwen3.5-9B & 0.099 & 0.155 & 0.525 & 0.670 & 0.426 & 0.424 & 0.383 \\
\bottomrule
\end{tabular}
\end{table}

\paragraph{The trained agents are competitive with larger models.}
Table~\ref{tab:leaderboard} places both trained agents among substantially larger systems, and where each one gains ground mirrors the route comparison above. The 4B agent ranks ninth, ahead of Gemma-4-31B-IT and Qwen3.5-9B and nearly tied with Qwen3.5-122B-A10B and Qwen3.6-35B-A3B. Its edge over these larger Qwen models lies in F1, which is higher under every judge, whereas its rubric scores under the independent judges remain lower, consistent with the F1 gain that the curriculum brings at 4B. The 27B agent ranks first, ahead of GPT-5.6-Luna and Kimi-K3, and its lead follows the curriculum's rubric advantage at this scale: it has the best rubric score under all three judges and the best \oss F1, while GPT-5.6-Luna keeps the best \gemma and \mistraljudge F1. This lead is largest on the training judge's rubric and narrower under the independent judges. Task-specific RL thus allows a 4B agent to match larger general-purpose models and a 27B agent to lead them on this task.

\subsection{Qualitative report comparison}
\label{sec:qualitative}
\afterpage{\begin{figure}[!t]
\centering
\begingroup\microtypesetup{expansion=false}
\setlength{\parindent}{0pt}%
\providecolor{PairSlate}{HTML}{475569}%
\tcbset{pairreport/.style args={#1}{enhanced,nobeforeafter,colback=white,
  colbacktitle=#1,colframe=#1,coltitle=white,
  fonttitle=\scriptsize\bfseries,boxrule=0.55pt,arc=1.2mm,
  left=3.5pt,right=3.5pt,top=1pt,bottom=1pt,
  toptitle=1pt,bottomtitle=1pt,boxsep=0pt}}%
\begin{tcolorbox}[pairreport=PairSlate,title={Curriculum RL report with GPT-OSS-120B claim labels}]
\fontsize{8.5}{9.4}\selectfont\setlength{\parskip}{1pt}%
\NclaimBg{1}{A longwave 500 mb trough currently sits over the northern Plains and Canadian prairies, with the upper-level jet core (250 mb winds near 65 m/s) overhead and a surface low centered near the northern Plains-Canadian border around 1008 hPa.} \NclaimBg{2}{A cold front extends south from that low across the central Plains, marking a sharp 850 mb thermal gradient between sub-freezing air over the northern Plains and warm air (8--12~°C at 850 mb) over the South.} \NclaimBg{3}{A secondary low over the southern Plains, near the Oklahoma-Kansas-Texas Panhandle area, is tapping warm, moist Gulf air and is the main source of heavy rain, with 12-hour accumulations already reaching 50--70 mm and 6-hour totals near 45 mm.}

\SclaimBg{1}{Over the next 24--48 hours the upper trough and its surface low will track east across the northern Plains and into the Upper Midwest and Great Lakes, bringing a broad band of showers and storms to the northern Plains, the Dakotas, and the Lakes through Friday into Saturday.} \SclaimBg{2}{Cooler air behind the front will cap daytime highs and allow snow showers along the northern Rockies and higher terrain where 850 mb temperatures remain below freezing.} \SclaimBg{3}{Meanwhile the southern Plains low and front will persist into the weekend, maintaining a risk of scattered to frequent thunderstorms with locally heavy rain, flash-flood potential, and the threat of severe weather given the warm, moist airmass and strong mid-level flow.}

\RclaimBg{1}{The Southeast and much of the South remain under the eastern flank of a 500 mb ridge, where conditions stay warm, stable, and largely dry, with 2 m temperatures in the low to mid 30s~°C and only isolated afternoon activity.} \NclaimBg{4}{The Pacific Northwest and northern Rockies will see intermittent precipitation from the western flank of the trough.} \SclaimBg{4}{The dominant hazard through the period is therefore the Plains frontal system: a moving band of precipitation and cool-air advection across the north, and a persistent, potentially severe and flood-prone storm track across the southern Plains, while the South enjoys warm, ridge-dominated weather.}
\end{tcolorbox}
\vspace{1pt}
\begin{tcolorbox}[pairreport=PairSlate,title={Reference report with evidence for SUPPORTED and REFUTED claims}]
\fontsize{8.5}{9.4}\selectfont\setlength{\parskip}{1pt}%
The meteorological discussion and forecast for the continental United States is: \EsentR{1}{Severe thunderstorms and heavy rainfall with the threat of flash flooding are expected for parts of the central and eastern U.S.}

A front stretched across the Ohio and Tennessee Valleys southwestward into the Lower Mississippi Valley and Southern Plains is expected to slowly make its way eastward across the eastern U.S. through the end of the week. \EsentBg{3}{Moisture from the Gulf of Mexico will cause widespread rain and thunderstorms along and ahead of the front.} \EsentBg{3}{Severe weather and heavy rain causing flash flooding are both possible.} \EsentBg{3}{On Friday, the most likely area for severe weather will be in southern Texas.} \EsentBg{3}{There is also a risk of excessive rainfall and flash flooding for much of Texas and into Louisiana and Arkansas.} Isolated strong thunderstorms and heavy rainfall with the possibility of causing flooding are possible into the Ohio Valley and Mid-Atlantic region as well. On Saturday, heavier rainfall totals are expected across the Ohio Valley, with a risk of flash flooding there. \EsentR{1}{The severe weather threat is forecast farther south across the Southeast and Southern Appalachians.}

The front will divide warmer than average temperatures to the east of it and cooler than average temperatures to the west. \EsentBg{4}{The Southeast and Tennessee Valley will be 5 to 10 degrees above average and the Mid-Atlantic region 10 to 15 degrees above average for high temperatures on Friday.} A backdoor front across the northern Mid-Atlantic on Friday will keep the Northeast cooler than average. Behind the main front, temperatures are forecast to be 5 to 15 degrees cooler than average in the Southern/Central Plains and into the Middle Mississippi Valley. As the front moves eastward on Saturday, the Ohio Valley could see high temperatures 5 to 10 degrees below average.

\EsentBg{3}{Farther west, scattered thunderstorms are forecast to develop on Saturday in the Southern and Central High Plains, which could be severe.} \EsentBg{1}{A cold front coming into the Northern Plains late Friday will progress into the Upper Midwest on Saturday, causing showers along it.} \EsentBg{2}{Temperatures are expected to cool below average in the Northern Rockies and Plains with this front, which could lead to light snow in those areas on Saturday.} The rest of the West will warm and have above average temperatures for the most part by Saturday, and generally remain dry ahead of a low approaching California by Sunday morning.
\end{tcolorbox}
\caption{Terminal \qwenL \Curr report with every \oss claim labeled SUPPORTED (\textcolor{ClaimSup}{\textbf{S}}), REFUTED (\textcolor{ClaimRef}{\textbf{R}}), or NEUTRAL (\textbf{N}). SUPPORTED and REFUTED claims are shaded green and red, and the author-mapped reference sentences that support or refute them carry the same tag and shading.}
\label{fig:reportviz}
\endgroup
\end{figure}
}Figure~\ref{fig:reportviz} shows the report from the terminal step-163 27B curriculum-trained checkpoint, together with the reference sentences that support each claim \oss labels SUPPORTED. The supported claims track the northern system and its showers into the Upper Midwest, link the trailing cold air to high-terrain snow, and identify the southern-Plains severe-weather and flash-flood threat. The reference has a counterpart for each, although it describes the northern system as a cold front and does not mention the upper trough or surface low. The report also makes a consequential regional error by describing the Southeast and much of the South as stable and largely dry despite the reference forecast of eastern-U.S. rainfall and a Southeast severe-weather threat. This sentence is the one claim that \oss labels REFUTED. The summary claim S4 repeats the error in its final clause, yet \oss labels it SUPPORTED, whereas \mistraljudge refutes that clause separately. Overall, the policy produces a coherent broad-scale account but can still introduce a locally important contradiction. Appendix~\ref{app:report-trajectory} traces the tool-based evidence behind this report, and Appendix~\ref{app:judgeviz} shows its claim labels and reference evidence under the two independent judges.

\section{Discussion}
\label{sec:conclusion}
In this work, we trained LLM-based agents to generate weather reports. We designed a reward that combines two LLM-as-a-judge metrics, a meteorological rubric and Discussion F1, and audited the judges' claim labels against three meteorologists' labels. Our two-stage curriculum RL transfers skills learned on simple, verifiable weather-analysis questions to judge-scored report generation. At the terminal checkpoint, \Curr on \qwenS improves F1 over the base policy and \Direct under all three judges, and on \qwenL it improves every rubric and F1 score over the base policy. More broadly, our results suggest that non-verifiable scientific tasks can still benefit from RL when careful reward design is combined with curricula that bootstrap from verifiable intermediate skills.

\textbf{Limitations and future work.} Our experiments focus on forecast discussions for the continental United States because of data availability, but the framework is not location-specific and can be extended to other regions and forecast products. The agent also receives oracle forecast inputs, and coupling it with operational numerical forecasts would test its end-to-end skill. Our evaluation relies on LLM judges. We use three judges and audit their claim labels against meteorologists' labels (Section~\ref{sec:humans}); evaluating expert preference over complete reports is an important next step. Finally, each configuration is trained once, and repeated training runs would quantify optimization variance.

\normalsize
\section*{Acknowledgments}
This work was supported in part by the U.S. Army Research Office under Army-ECASE award W911NF-07-R-0003-03, the U.S. Department of Energy, Office of Science, the ARPA-H-SOL-24-101 program, the IARPA HAYSTAC Program, NSF Grants \#2205093, \#2146343, \#2134274, \#2441832, \#2146151, and \#2422214, NSF award CCF-2112665 (TILOS), and CDC-RFA-FT-23-0069. We also thank Thinking Machines Lab for providing Tinker compute credits.

\bibliography{references}

\begin{thebibliography}{51}
\providecommand{\natexlab}[1]{#1}
\providecommand{\url}[1]{\texttt{#1}}
\expandafter\ifx\csname urlstyle\endcsname\relax
  \providecommand{\doi}[1]{doi: #1}\else
  \providecommand{\doi}{doi: \begingroup \urlstyle{rm}\Url}\fi

\bibitem[Agarwal et~al.(2025)Agarwal, Ahmad, Ai, Altman, Applebaum, Arbus,
  Arora, Bai, Baker, Bao, et~al.]{agarwal2025gptoss}
Sandhini Agarwal, Lama Ahmad, Jason Ai, Sam Altman, Andy Applebaum, Edwin
  Arbus, Rahul~K. Arora, Yu~Bai, Bowen Baker, Haiming Bao, et~al.
\newblock gpt-oss-120b \& gpt-oss-20b model card.
\newblock arXiv preprint arXiv:2508.10925, 2025.

\bibitem[Alet et~al.(2026)Alet, Andersson, Price, Markou, El-Kadi, Masters, Li,
  Merchant, Williams, Thornton, MacKay, Graham, Uddin, Gaiarin, Shah, Kruse,
  Hogsett, Zelinsky, Cangialosi, Martinez, Franklin, DeMaria, Musgrave, Bain,
  Titley, Stott, Lam, Bell, Komarek, Willson, Sanchez-Gonzalez, and
  Battaglia]{Alet2026}
Ferran Alet, Tom~R. Andersson, Ilan Price, Stratis Markou, Andrew El-Kadi,
  Dominic Masters, Amy Li, Samier Merchant, Natalie Williams, Gregory Thornton,
  Ken MacKay, Olivia Graham, Akib Uddin, Ben Gaiarin, Devaja Shah, Elinor
  Kruse, Wallace Hogsett, David Zelinsky, John Cangialosi, Jonathan Martinez,
  James Franklin, Mark DeMaria, Kate Musgrave, Caroline~L. Bain, Helen Titley,
  Jacklynn Stott, Remi Lam, Aaron Bell, Paul Komarek, Matthew Willson, Alvaro
  Sanchez-Gonzalez, and Peter Battaglia.
\newblock Operational tropical cyclone forecasting with {AI}.
\newblock \emph{Nature}, 657\penalty0 (8132):\penalty0 680--688, 2026.

\bibitem[Bengio et~al.(2009)Bengio, Louradour, Collobert, and
  Weston]{bengio2009curriculum}
Yoshua Bengio, J{\'e}r{\^o}me Louradour, Ronan Collobert, and Jason Weston.
\newblock Curriculum learning.
\newblock In \emph{Proceedings of the 26th Annual International Conference on
  Machine Learning}, pages 41--48, 2009.

\bibitem[Boiko et~al.(2023)Boiko, MacKnight, Kline, and
  Gomes]{boiko2023coscientist}
Daniil~A. Boiko, Robert MacKnight, Ben Kline, and Gabe Gomes.
\newblock Autonomous chemical research with large language models.
\newblock \emph{Nature}, 624\penalty0 (7992):\penalty0 570--578, 2023.

\bibitem[Bran et~al.(2024)Bran, Cox, Schilter, Baldassari, White, and
  Schwaller]{bran2024chemcrow}
Andres~M. Bran, Sam Cox, Oliver Schilter, Carlo Baldassari, Andrew~D. White,
  and Philippe Schwaller.
\newblock Augmenting large language models with chemistry tools.
\newblock \emph{Nature Machine Intelligence}, 6\penalty0 (5):\penalty0
  525--535, 2024.

\bibitem[Chen et~al.(2025)Chen, Lu, Kim, Zhang, Tang, Pich{\'e}, Gontier,
  Bengio, and Kamalloo]{chen2025selfevolving}
Xiaoyin Chen, Jiarui Lu, Minsu Kim, Dinghuai Zhang, Jian Tang, Alexandre
  Pich{\'e}, Nicolas Gontier, Yoshua Bengio, and Ehsan Kamalloo.
\newblock Self-evolving curriculum for {LLM} reasoning.
\newblock arXiv preprint arXiv:2505.14970, 2025.

\bibitem[{DeepSeek-AI}(2026)]{deepseek2026vision}
{DeepSeek-AI}.
\newblock {DeepSeek-V4-Flash-Vision-Exp} release: Multimodal {API} now live.
\newblock \url{https://api-docs.deepseek.com/news/news260821/}, August 2026.

\bibitem[Feng et~al.(2025)Feng, Huang, Qu, Zhang, Qin, Zhong, Jiang, Chi, and
  Zhong]{feng2025retool}
Jiazhan Feng, Shijue Huang, Xingwei Qu, Ge~Zhang, Yujia Qin, Baoquan Zhong,
  Chengquan Jiang, Jinxin Chi, and Wanjun Zhong.
\newblock {ReTool}: Reinforcement learning for strategic tool use in {LLMs}.
\newblock arXiv preprint arXiv:2504.11536, 2025.

\bibitem[{Gemma Team} et~al.(2025){Gemma Team}, Kamath, Ferret, Pathak,
  Vieillard, Merhej, Perrin, Matejovicova, Ram{\'e}, Rivi{\`e}re, Rouillard,
  et~al.]{gemmateam2025gemma3}
{Gemma Team}, Aishwarya Kamath, Johan Ferret, Shreya Pathak, Nino Vieillard,
  Ramona Merhej, Sarah Perrin, Tatiana Matejovicova, Alexandre Ram{\'e},
  Morgane Rivi{\`e}re, Louis Rouillard, et~al.
\newblock Gemma 3 technical report.
\newblock arXiv preprint arXiv:2503.19786, 2025.

\bibitem[{Gemma Team} et~al.(2026){Gemma Team}, Abd, Aggarwal, Algayres,
  Andreev, Bachem, Ballantyne, Brick, C{\u{a}}rbune, Casbon, Chaturvedi,
  et~al.]{gemmateam2026gemma4}
{Gemma Team}, Sherif~El Abd, Vaibhav Aggarwal, Robin Algayres, Alek Andreev,
  Olivier Bachem, Ian Ballantyne, Cormac Brick, Victor C{\u{a}}rbune, Michelle
  Casbon, Mayank Chaturvedi, et~al.
\newblock Gemma 4 technical report.
\newblock arXiv preprint arXiv:2607.02770, 2026.

\bibitem[{GLM-5-Team} et~al.(2026){GLM-5-Team}, Zeng, Lv, Hou, Du, Zheng, Chen,
  Yin, Ge, Huang, Xie, et~al.]{zhipu2026glm5}
{GLM-5-Team}, Aohan Zeng, Xin Lv, Zhenyu Hou, Zhengxiao Du, Qinkai Zheng, Bin
  Chen, Da~Yin, Chendi Ge, Chenghua Huang, Chengxing Xie, et~al.
\newblock {GLM-5}: from vibe coding to agentic engineering.
\newblock arXiv preprint arXiv:2602.15763, 2026.

\bibitem[Gunjal et~al.(2025)Gunjal, Wang, Lau, Nath, He, Liu, and
  Hendryx]{gunjal2025rubrics}
Anisha Gunjal, Anthony Wang, Elaine Lau, Vaskar Nath, Yunzhong He, Bing Liu,
  and Sean Hendryx.
\newblock Rubrics as rewards: Reinforcement learning beyond verifiable domains.
\newblock arXiv preprint arXiv:2507.17746, 2025.

\bibitem[Guo et~al.(2025)Guo, Yang, Zhang, Song, Wang, Zhu, Xu, Zhang,
  et~al.]{guo2025deepseekr1nature}
Daya Guo, Dejian Yang, Haowei Zhang, Junxiao Song, Peiyi Wang, Qihao Zhu,
  Runxin Xu, Ruoyu Zhang, et~al.
\newblock {DeepSeek-R1} incentivizes reasoning in {LLMs} through reinforcement
  learning.
\newblock \emph{Nature}, 645\penalty0 (8081):\penalty0 633--638, 2025.

\bibitem[Hao et~al.(2026)Hao, Jin, Liao, Liu, and Zhao]{hao2026toolcollapse}
Yupu Hao, Zhuoran Jin, Huanxuan Liao, Kang Liu, and Jun Zhao.
\newblock Why multi-step tool-use reinforcement learning collapses and how
  supervisory signals fix it.
\newblock arXiv preprint arXiv:2606.26027, 2026.

\bibitem[Hersbach et~al.(2020)Hersbach, Bell, Berrisford, Hirahara,
  Hor{\'a}nyi, Mu{\~n}oz-Sabater, Nicolas, Peubey, Radu, Schepers, Simmons,
  Soci, Abdalla, Abellan, Balsamo, Bechtold, Biavati, Bidlot, Bonavita, Chiara,
  Dahlgren, Dee, Diamantakis, Dragani, Flemming, Forbes, Fuentes, Geer,
  Haimberger, Healy, Hogan, H{\'o}lm, Janiskov{\'a}, Keeley, Laloyaux, Lopez,
  Lupu, Radnoti, de~Rosnay, Rozum, Vamborg, Villaume, and
  Th{\'e}paut]{hersbach2020era5}
Hans Hersbach, Bill Bell, Paul Berrisford, Shoji Hirahara, Andr{\'a}s
  Hor{\'a}nyi, Joaqu{\'i}n Mu{\~n}oz-Sabater, Julien Nicolas, Carole Peubey,
  Raluca Radu, Dinand Schepers, Adrian Simmons, Cornel Soci, Saleh Abdalla,
  Xavier Abellan, Gianpaolo Balsamo, Peter Bechtold, Gionata Biavati, Jean
  Bidlot, Massimo Bonavita, Giovanna~De Chiara, Per Dahlgren, Dick Dee, Michail
  Diamantakis, Rossana Dragani, Johannes Flemming, Richard Forbes, Manuel
  Fuentes, Alan Geer, Leo Haimberger, Sean Healy, Robin~J. Hogan, El{\'i}as
  H{\'o}lm, Marta Janiskov{\'a}, Sarah Keeley, Patrick Laloyaux, Philippe
  Lopez, Cristina Lupu, Gabor Radnoti, Patricia de~Rosnay, Iryna Rozum, Freja
  Vamborg, Sebastien Villaume, and Jean-No{\"e}l Th{\'e}paut.
\newblock The {ERA5} global reanalysis.
\newblock \emph{Quarterly Journal of the Royal Meteorological Society},
  146\penalty0 (730):\penalty0 1999--2049, 2020.

\bibitem[Higgins et~al.(2026)Higgins, Mamalakis, and
  Agarwal]{higgins2026synopticbench}
Timothy~B. Higgins, Antonios Mamalakis, and Chirag Agarwal.
\newblock {SynopticBench}: Evaluating vision-language models on generating
  weather forecast discussions of the future.
\newblock arXiv preprint arXiv:2604.16451, 2026.

\bibitem[Hu et~al.(2022)Hu, Shen, Wallis, Allen-Zhu, Li, Wang, Wang, and
  Chen]{hu2022lora}
Edward~J. Hu, Yelong Shen, Phillip Wallis, Zeyuan Allen-Zhu, Yuanzhi Li, Shean
  Wang, Lu~Wang, and Weizhu Chen.
\newblock {LoRA}: Low-rank adaptation of large language models.
\newblock In \emph{International Conference on Learning Representations}, 2022.

\bibitem[Huang et~al.(2025)Huang, Zhang, Wang, Qu, Lu, Roohani, Li, Qiu, Li,
  Zhang, Yin, Marwaha, Carter, Zhou, Wheeler, Bernstein, Wang, He, Zhou,
  Snyder, Cong, Regev, and Leskovec]{huang2025biomni}
Kexin Huang, Serena Zhang, Hanchen Wang, Yuanhao Qu, Yingzhou Lu, Yusuf
  Roohani, Ryan Li, Lin Qiu, Gavin Li, Junze Zhang, Di~Yin, Shruti Marwaha,
  Jennefer~N. Carter, Xin Zhou, Matthew Wheeler, Jonathan~A. Bernstein, Mengdi
  Wang, Peng He, Jingtian Zhou, Michael Snyder, Le~Cong, Aviv Regev, and Jure
  Leskovec.
\newblock Biomni: A general-purpose biomedical {AI} agent.
\newblock bioRxiv preprint, 2025.

\bibitem[Jiang et~al.(2026)Jiang, Wang, Zhang, Zeng, Li, Chen, Tao, Bai, and
  Shang]{jiang2026rlvrr}
Yuxin Jiang, Yufei Wang, Qiyuan Zhang, Xingshan Zeng, Liangyou Li, Jierun Chen,
  Chaofan Tao, Haoli Bai, and Lifeng Shang.
\newblock From verifiable dot to reward chain: Harnessing verifiable
  reference-based rewards for reinforcement learning of open-ended generation.
\newblock In \emph{The Fourteenth International Conference on Learning
  Representations}, 2026.

\bibitem[Jin et~al.(2025)Jin, Zeng, Yue, Yoon, Arik, Wang, Zamani, and
  Han]{jin2025searchr1}
Bowen Jin, Hansi Zeng, Zhenrui Yue, Jinsung Yoon, Sercan Arik, Dong Wang, Hamed
  Zamani, and Jiawei Han.
\newblock {Search-R1}: Training {LLMs} to reason and leverage search engines
  with reinforcement learning.
\newblock arXiv preprint arXiv:2503.09516, 2025.

\bibitem[{Kimi Team} et~al.(2026){Kimi Team}, Bai, Bai, Bao, C., Cai, Cai, Cao,
  Cao, Chai, et~al.]{moonshot2026kimik3}
{Kimi Team}, Tongtong Bai, Yifan Bai, Yiping Bao, M.~C., Jianfeng Cai, Xinyuan
  Cai, Peizhou Cao, Yuxuan Cao, Ziwei Chai, et~al.
\newblock {Kimi K3}: Open frontier intelligence.
\newblock arXiv preprint arXiv:2607.24653, 2026.

\bibitem[Lai et~al.(2026)Lai, Xu, Yang, Chen, Xu, Zeng, Li, Sun, Zhu, Zhang,
  Hu, Li, Gao, Li, Zhu, Zhou, and Zhao]{minimax2026msa}
Xunhao Lai, Weiqi Xu, Yufeng Yang, Qiaorui Chen, Yang Xu, Lunbin Zeng, Xiaolong
  Li, Haohai Sun, Haichao Zhu, Vito Zhang, Jinkai Hu, Jiayao Li, Rui Gao, Zekun
  Li, Songquan Zhu, Jingkai Zhou, and Pengyu Zhao.
\newblock {MiniMax} sparse attention.
\newblock arXiv preprint arXiv:2606.13392, 2026.

\bibitem[Lawson et~al.(2025)Lawson, Trujillo-Falc{\'o}n, Schultz, Flora,
  Goebbert, Lyman, Potvin, and Stepanek]{lawson2025pixels}
John~R Lawson, Joseph~E Trujillo-Falc{\'o}n, David~M Schultz, Montgomery~L
  Flora, Kevin~H Goebbert, Seth~N Lyman, Corey~K Potvin, and Adam~J Stepanek.
\newblock Pixels and predictions: Potential of {GPT-4V} in meteorological
  imagery analysis and forecast communication.
\newblock \emph{Artificial Intelligence for the Earth Systems}, 4\penalty0 (1),
  2025.

\bibitem[Li et~al.(2025)Li, Zhang, Yin, Zhang, Ou, Wu, Yin, Li, Tao, Wang,
  Shen, Zhang, Zhang, Wu, Jiang, Yan, Xie, Huang, and Zhou]{li2025websailor}
Kuan Li, Zhongwang Zhang, Huifeng Yin, Liwen Zhang, Litu Ou, Jialong Wu,
  Wenbiao Yin, Baixuan Li, Zhengwei Tao, Xinyu Wang, Weizhou Shen, Junkai
  Zhang, Dingchu Zhang, Xixi Wu, Yong Jiang, Ming Yan, Pengjun Xie, Fei Huang,
  and Jingren Zhou.
\newblock {WebSailor}: Navigating super-human reasoning for web agent.
\newblock arXiv preprint arXiv:2507.02592, 2025.

\bibitem[{MiniMax}(2025)]{minimax2025m1}
{MiniMax}.
\newblock {MiniMax-M1}: Scaling test-time compute efficiently with lightning
  attention.
\newblock arXiv preprint arXiv:2506.13585, 2025.

\bibitem[{Mistral AI}(2025)]{mistral2025small32}
{Mistral AI}.
\newblock {Mistral Small 3.2 (24B Instruct 2506)}.
\newblock
  \url{https://huggingface.co/mistralai/Mistral-Small-3.2-24B-Instruct-2506},
  June 2025.

\bibitem[Narvekar et~al.(2020)Narvekar, Peng, Leonetti, Sinapov, Taylor, and
  Stone]{narvekar2020curriculum}
Sanmit Narvekar, Bei Peng, Matteo Leonetti, Jivko Sinapov, Matthew~E. Taylor,
  and Peter Stone.
\newblock Curriculum learning for reinforcement learning domains: A framework
  and survey.
\newblock \emph{Journal of Machine Learning Research}, 21\penalty0
  (181):\penalty0 1--50, 2020.

\bibitem[{OpenAI}(2026)]{openai2026gpt56}
{OpenAI}.
\newblock {GPT-5.6} system card.
\newblock \url{https://deploymentsafety.openai.com/gpt-5-6}, 2026.

\bibitem[Ouyang et~al.(2022)Ouyang, Wu, Jiang, Almeida, Wainwright, Mishkin,
  Zhang, Agarwal, Slama, Ray, Schulman, Hilton, Kelton, Miller, Simens, Askell,
  Welinder, Christiano, Leike, and Lowe]{ouyang2022instructgpt}
Long Ouyang, Jeff Wu, Xu~Jiang, Diogo Almeida, Carroll~L. Wainwright, Pamela
  Mishkin, Chong Zhang, Sandhini Agarwal, Katarina Slama, Alex Ray, John
  Schulman, Jacob Hilton, Fraser Kelton, Luke Miller, Maddie Simens, Amanda
  Askell, Peter Welinder, Paul Christiano, Jan Leike, and Ryan Lowe.
\newblock Training language models to follow instructions with human feedback.
\newblock In \emph{Advances in Neural Information Processing Systems},
  volume~35, 2022.

\bibitem[Parashar et~al.(2025)Parashar, Gui, Li, Ling, Vemuri, Olson, Li,
  Zhang, Caverlee, Kalathil, and Ji]{parashar2025e2h}
Shubham Parashar, Shurui Gui, Xiner Li, Hongyi Ling, Sushil Vemuri, Blake
  Olson, Eric Li, Yu~Zhang, James Caverlee, Dileep Kalathil, and Shuiwang Ji.
\newblock Curriculum reinforcement learning from easy to hard tasks improves
  {LLM} reasoning.
\newblock arXiv preprint arXiv:2506.06632, 2025.

\bibitem[{Qwen Team}(2026{\natexlab{a}})]{qwenteam2026qwen35}
{Qwen Team}.
\newblock {Qwen3.5}: Towards native multimodal agents.
\newblock \url{https://qwen.ai/blog?id=qwen3.5}, February 2026{\natexlab{a}}.

\bibitem[{Qwen Team}(2026{\natexlab{b}})]{qwenteam2026qwen36}
{Qwen Team}.
\newblock {Qwen3.6-35B-A3B}: Agentic coding power, now open to all.
\newblock \url{https://qwen.ai/blog?id=qwen3.6-35b-a3b}, April
  2026{\natexlab{b}}.

\bibitem[{Qwen Team}(2026{\natexlab{c}})]{qwenteam2026qwen38}
{Qwen Team}.
\newblock {Qwen3.8-27B} model card.
\newblock \url{https://huggingface.co/Qwen/Qwen3.8-27B}, August
  2026{\natexlab{c}}.

\bibitem[Rasp et~al.(2024)Rasp, Hoyer, Merose, Langmore, Battaglia, Russell,
  Sanchez-Gonzalez, Yang, Carver, Agrawal, Chantry, Bouallegue, Dueben,
  Bromberg, Sisk, Barrington, Bell, and Sha]{rasp2024weatherbench2}
Stephan Rasp, Stephan Hoyer, Alexander Merose, Ian Langmore, Peter Battaglia,
  Tyler Russell, Alvaro Sanchez-Gonzalez, Vivian Yang, Rob Carver, Shreya
  Agrawal, Matthew Chantry, Zied~Ben Bouallegue, Peter Dueben, Carla Bromberg,
  Jared Sisk, Luke Barrington, Aaron Bell, and Fei Sha.
\newblock {WeatherBench} 2: A benchmark for the next generation of data-driven
  global weather models.
\newblock \emph{Journal of Advances in Modeling Earth Systems}, 16\penalty0
  (6):\penalty0 e2023MS004019, 2024.

\bibitem[Schick et~al.(2023)Schick, Dwivedi-Yu, Dess{\`i}, Raileanu, Lomeli,
  Hambro, Zettlemoyer, Cancedda, and Scialom]{schick2023toolformer}
Timo Schick, Jane Dwivedi-Yu, Roberto Dess{\`i}, Roberta Raileanu, Maria
  Lomeli, Eric Hambro, Luke Zettlemoyer, Nicola Cancedda, and Thomas Scialom.
\newblock Toolformer: Language models can teach themselves to use tools.
\newblock In \emph{Advances in Neural Information Processing Systems},
  volume~36, 2023.

\bibitem[Shao et~al.(2024)Shao, Wang, Zhu, Xu, Song, Bi, Zhang, Zhang, Li, Wu,
  and Guo]{shao2024deepseekmath}
Zhihong Shao, Peiyi Wang, Qihao Zhu, Runxin Xu, Junxiao Song, Xiao Bi, Haowei
  Zhang, Mingchuan Zhang, Y.~K. Li, Y.~Wu, and Daya Guo.
\newblock {DeepSeekMath}: Pushing the limits of mathematical reasoning in open
  language models.
\newblock arXiv preprint arXiv:2402.03300, 2024.

\bibitem[Shi et~al.(2026)Shi, Wu, Song, Zhou, and Zhao]{shi2025adarft}
Taiwei Shi, Yiyang Wu, Linxin Song, Tianyi Zhou, and Jieyu Zhao.
\newblock Efficient reinforcement finetuning via adaptive curriculum learning.
\newblock \emph{Transactions on Machine Learning Research}, 2026.

\bibitem[Shinn et~al.(2023)Shinn, Cassano, Gopinath, Narasimhan, and
  Yao]{shinn2023reflexion}
Noah Shinn, Federico Cassano, Ashwin Gopinath, Karthik Narasimhan, and Shunyu
  Yao.
\newblock Reflexion: Language agents with verbal reinforcement learning.
\newblock In \emph{Advances in Neural Information Processing Systems},
  volume~36, 2023.

\bibitem[Singh et~al.(2026)Singh, Luitel, Singh, Banga, Sudharsan, and
  Durkee]{singh2026afdbench}
Manmeet Singh, Somnath Luitel, Prabhjot Singh, Manraaj Banga, Naveen Sudharsan,
  and Josh Durkee.
\newblock {AFDBench}: A reasoning-first {AI} scientist for {National Weather
  Service} forecast discussions.
\newblock arXiv preprint arXiv:2608.24954, 2026.

\bibitem[Tao et~al.(2025)Tao, Kulikov, Saha, Wang, Xu, Li, Weston, and
  Yu]{tao2025hero}
Leitian Tao, Ilia Kulikov, Swarnadeep Saha, Tianlu Wang, Jing Xu, Sharon Li,
  Jason~E. Weston, and Ping Yu.
\newblock Hybrid reinforcement: When reward is sparse, it's better to be dense.
\newblock arXiv preprint arXiv:2510.07242, 2025.

\bibitem[Thaman(2026)]{thaman2026rewardhacking}
Kunvar Thaman.
\newblock Reward hacking benchmark: Measuring exploits in {LLM} agents with
  tool use.
\newblock arXiv preprint arXiv:2605.02964, 2026.

\bibitem[Ullrich and Zarzycki(2017)]{ullrich2017tempestextremes}
Paul~A. Ullrich and Colin~M. Zarzycki.
\newblock {TempestExtremes}: A framework for scale-insensitive pointwise
  feature tracking on unstructured grids.
\newblock \emph{Geoscientific Model Development}, 10\penalty0 (3):\penalty0
  1069--1090, 2017.

\bibitem[Varambally et~al.(2026)Varambally, Fisher, Thakker, Chen, Xia, Jafari,
  Niu, Jain, Manivannan, Novack, Han, Eranky, Cachay, Berg-Kirkpatrick,
  Watson-Parris, Ma, and Yu]{varambally2026zephyrus}
Sumanth Varambally, Marshall Fisher, Jas Thakker, Yiwei Chen, Zhirui Xia,
  Yasaman Jafari, Ruijia Niu, Manas Jain, Veeramakali~Vignesh Manivannan,
  Zachary Novack, Luyu Han, Srikar Eranky, Salva~R{\"u}hling Cachay, Taylor
  Berg-Kirkpatrick, Duncan Watson-Parris, Yian Ma, and Rose Yu.
\newblock Zephyrus: An agentic framework for weather science.
\newblock In \emph{The Fourteenth International Conference on Learning
  Representations}, 2026.

\bibitem[Viswanathan et~al.(2025)Viswanathan, Sun, Kong, Cao, Neubig, and
  Wu]{viswanathan2025checklists}
Vijay Viswanathan, Yanchao Sun, Xiang Kong, Meng Cao, Graham Neubig, and Sherry
  Wu.
\newblock Checklists are better than reward models for aligning language
  models.
\newblock In \emph{Advances in Neural Information Processing Systems},
  volume~38, 2025.

\bibitem[Wei et~al.(2026)Wei, Zong, Li, Yu, and Li]{wei2026qurl}
Xiyu Wei, Qingwei Zong, Xiaoguang Li, Eugene Yu, and Sujian Li.
\newblock {QuRL}: Rubrics as judge for open-ended question answering.
\newblock In \emph{The Fourteenth International Conference on Learning
  Representations}, 2026.

\bibitem[Yang et~al.(2025)Yang, Li, Yang, Zhang, Hui, Zheng, Yu, Gao, Huang,
  Lv, et~al.]{yang2025qwen3}
An~Yang, Anfeng Li, Baosong Yang, Beichen Zhang, Binyuan Hui, Bo~Zheng, Bowen
  Yu, Chang Gao, Chengen Huang, Chenxu Lv, et~al.
\newblock Qwen3 technical report.
\newblock arXiv preprint arXiv:2505.09388, 2025.

\bibitem[Yao et~al.(2023)Yao, Zhao, Yu, Du, Shafran, Narasimhan, and
  Cao]{yao2023react}
Shunyu Yao, Jeffrey Zhao, Dian Yu, Nan Du, Izhak Shafran, Karthik Narasimhan,
  and Yuan Cao.
\newblock {ReAct}: Synergizing reasoning and acting in language models.
\newblock In \emph{The Eleventh International Conference on Learning
  Representations}, 2023.

\bibitem[Yu et~al.(2025)Yu, Zhang, Zhu, Yuan, Zuo, Yue, Dai, Fan, Liu, Liu,
  Liu, Liu, Lin, Lin, Ma, Sheng, Tong, Zhang, Zhang, Zhang, Zhang, Zhu, Zhu,
  Chen, Chen, Wang, Yu, Song, Wei, Zhou, Liu, Ma, Zhang, Yan, Wu, and
  Wang]{yu2025dapo}
Qiying Yu, Zheng Zhang, Ruofei Zhu, Yufeng Yuan, Xiaochen Zuo, Yu~Yue, Weinan
  Dai, Tiantian Fan, Gaohong Liu, Juncai Liu, Lingjun Liu, Xin Liu, Haibin Lin,
  Zhiqi Lin, Bole Ma, Guangming Sheng, Yuxuan Tong, Chi Zhang, Mofan Zhang,
  Ru~Zhang, Wang Zhang, Hang Zhu, Jinhua Zhu, Jiaze Chen, Jiangjie Chen,
  Chengyi Wang, Hongli Yu, Yuxuan Song, Xiangpeng Wei, Hao Zhou, Jingjing Liu,
  Wei-Ying Ma, Ya-Qin Zhang, Lin Yan, Yonghui Wu, and Mingxuan Wang.
\newblock {DAPO}: An open-source {LLM} reinforcement learning system at scale.
\newblock In \emph{Advances in Neural Information Processing Systems},
  volume~38, 2025.

\bibitem[Zheng et~al.(2023)Zheng, Chiang, Sheng, Zhuang, Wu, Zhuang, Lin, Li,
  Li, Xing, Zhang, Gonzalez, and Stoica]{zheng2023judging}
Lianmin Zheng, Wei-Lin Chiang, Ying Sheng, Siyuan Zhuang, Zhanghao Wu, Yonghao
  Zhuang, Zi~Lin, Zhuohan Li, Dacheng Li, Eric~P. Xing, Hao Zhang, Joseph~E.
  Gonzalez, and Ion Stoica.
\newblock Judging {LLM}-as-a-judge with {MT-Bench} and {Chatbot Arena}.
\newblock In \emph{Advances in Neural Information Processing Systems (Datasets
  and Benchmarks Track)}, volume~36, 2023.

\bibitem[Zheng et~al.(2025)Zheng, Fu, Hu, Cai, Ye, Lu, and
  Liu]{zheng2025deepresearcher}
Yuxiang Zheng, Dayuan Fu, Xiangkun Hu, Xiaojie Cai, Lyumanshan Ye, Pengrui Lu,
  and Pengfei Liu.
\newblock {DeepResearcher}: Scaling deep research via reinforcement learning in
  real-world environments.
\newblock arXiv preprint arXiv:2504.03160, 2025.

\bibitem[Zheng et~al.(2026)Zheng, Liu, Chen, Zheng, Cheng, and
  Li]{zheng2026weathersyn}
Zinan Zheng, Yang Liu, Nuo Chen, Juepeng Zheng, Hong Cheng, and Jia Li.
\newblock {WeatherSyn}: An instruction tuning {MLLM} for weather forecasting
  report generation.
\newblock arXiv preprint arXiv:2605.07522, 2026.

\end{thebibliography}
\bibliographystyle{plainnat}

\clearpage
\appendix
\raggedbottom
\counterwithin{figure}{section}\counterwithin{table}{section}
\renewcommand{\thefigure}{\Alph{section}.\arabic{figure}}\renewcommand{\thetable}{\Alph{section}.\arabic{table}}
\section{Data}
\label{app:data}

\paragraph{Input fields.}
The \wb fields span 1959 to 2023 at six-hour intervals on a one-degree equiangular grid. Each event contains surface pressure and temperature fields, winds, precipitation, land and ocean fields, and pressure-level geopotential, humidity, temperature, winds, and vertical velocity.
The prompt exposes relative time indices and names the month, day, and weekday of the analysis window and the following five days because operational forecast discussions refer to days by name.

\paragraph{Report preprocessing and split checks.}
A fixed prompt removes author notes, map and model references, and disclaimers from each reference report. The mean reference lengths are approximately 1,800 characters in the training set and 2,000 in the test set. No test reference matches a training reference or shares its first 200 characters, and the test set contains no duplicate references. The prompt omits the year, so a dated event is not a unique historical identifier. The grounding requirement forces the agent to query the data.

\section{Stage-1 training and evaluation}
\label{app:curriculum}
Stage 1 trains on verifiable questions with rejection-sampling fine-tuning (RSFT), RL, or RSFT followed by RL. Table~\ref{tab:reward-constants} lists the rejection-sampling and RL parameters of both stages. The following subsections give the RSFT selection rules, the Stage-1 verifier, the RL settings and evaluation protocol, and example trajectories.

\subsection{Rejection-sampling fine-tuning}
\label{app:stage1-rft}
A rollout is eligible for RSFT if the answer receives full verifier credit, the server records at least one call to the question's required tool, no infrastructure failure occurs, the final \texttt{<solution>} block contains a valid JSON answer with finite values, and no turn omits both the \texttt{<execute>} and \texttt{<solution>} tags. A trajectory remains eligible if the model repairs an error in its own code.
RSFT keeps at most two eligible trajectories per question for each model, preferring fewer execution errors, fewer turns, and shorter responses, in that order. For 27B, trajectories that avoid a forced final turn receive an additional preference. The sample index breaks any remaining tie.
RSFT trains a rank-64 LoRA adapter on the language layers for three epochs with learning rate $10^{-5}$ and batch size four.
The 4B model produces 1{,}801 full-credit rollouts with required-tool access (31.7\%), compared with 3{,}125 (55.0\%) for 27B. The larger base policy therefore solves the verifiable tasks more reliably before fine-tuning. The two-per-question cap keeps repeated successes from dominating either RSFT corpus.

\subsection{Stage-1 verifier}
\label{app:verifier}
\begin{table}[!htb]
\caption{The eight \zephbench question templates used in Stage 1 and the tool required by each template.}
\label{tab:stage1-tasks}
\centering
\apptab
\begin{tabular*}{\linewidth}{@{\extracolsep{\fill}}L{0.64\linewidth}L{0.15\linewidth}L{0.14\linewidth}@{}}
\toprule
\textbf{Question template} & \textbf{Answer} & \textbf{Required tool} \\
\midrule
Geographic feature with the highest or lowest time-mean field value & location & dataset \\
\rowsep
Minimum, maximum, mean, or median field value over a named geographic feature & numerical & dataset \\
\rowsep
Sublocation within a named parent feature with the highest or lowest recorded field value & location & dataset \\
\rowsep
Hours from the initial time at which a named feature reaches its maximum or minimum field value & time & dataset \\
\rowsep
Minimum, maximum, mean, or median field value over a named feature at a specified time & numerical & dataset \\
\rowsep
Geographic features above or below a specified climatological percentile & anomaly list & climatology \\
\rowsep
Geographic features whose field differs from the climatological mean by a specified number of standard deviations & anomaly list & climatology \\
\rowsep
Geographic features outside a specified lower-upper climatological percentile envelope & anomaly list & climatology \\
\bottomrule
\end{tabular*}
\end{table}

\begin{table}[!htb]
\caption{Stage-1 answer formats and verifier scores. Numerical questions use absolute-error bands when the gold magnitude is below one. Anomaly lists receive graded set F1.}
\label{tab:verifier-ladder}
\centering
\apptab
\begin{tabular*}{\linewidth}{@{\extracolsep{\fill}}lL{0.20\linewidth}L{0.22\linewidth}L{0.20\linewidth}L{0.15\linewidth}@{}}
\toprule
\textbf{Type} & \textbf{Required content} & \textbf{Score 1.0} & \textbf{Score 0.5} & \textbf{Score 0} \\
\midrule
Numerical & one JSON number & relative error $\leq 1\%$ (absolute error $\leq 0.05$ if $|\mathrm{gold}|<1$) & relative error $\leq 5\%$ (absolute error $\leq 0.15$ if $|\mathrm{gold}|<1$) & otherwise \\
\rowsep
Time & one JSON number in hours & exact & within 6 h & otherwise \\
\rowsep
Location & JSON object with place name and value & normalized name equals or contains the gold name & wrong name, value inside the numerical 1.0 band & otherwise \\
\rowsep
Anomaly & JSON list of names (empty if none) & set F1 over normalized names, graded in $[0,1]$ & (graded) & no overlap \\
\bottomrule
\end{tabular*}
\end{table}

Stage 1 uses the eight \zephbench question templates in Table~\ref{tab:stage1-tasks}. The training and held-out validation questions are disjoint but follow the same templates. Answers may be numerical values, locations, times, or anomaly lists. Gold values come from the benchmark metadata. For location questions, we parse the gold place name from the reference response during data construction. The model returns JSON inside the \texttt{<solution>} block. Table~\ref{tab:verifier-ladder} gives the required format and score ladder. Numerical questions use relative error $|\mathrm{pred}-\mathrm{gold}|/\max(|\mathrm{gold}|,10^{-9})$, with absolute-error bands when $|\mathrm{gold}|<1$.

The ladder gives partial credit to nearby numerical or temporal values, correct values paired with the wrong location, and partially correct anomaly sets. A separate tool check prevents an unsupported guess from qualifying as a successful trajectory.

In Equation~\ref{eq:stage1}, $s$ is the ladder score and the turn penalty is $\lambda(T)=\min(0.15,\,0.05\max(0,T-5))$ for an episode with $T$ turns.
The required-tool condition holds when the server records at least one call to the dataset or climatology tool, whichever the question requires.

\subsection{Stage-1 RL and evaluation}
\label{app:stage1-val}
\paragraph{Settings.}
Stage-1 RL uses the report-stage trainer with learning rate $10^{-5}$ but replaces the judge-based reward with the accuracy reward in Equation~\ref{eq:stage1}. RSFT-RL starts from the selected RSFT checkpoint, while RL-only starts from the base policy. Each run takes 44 steps, drawing 704 questions from the 710-question pool. Report RL is initialized from RL-only step 44 for \qwenS and RSFT-RL step 24 for \qwenL.

\begin{figure}[!htb]
\centering
\includegraphics[width=\linewidth]{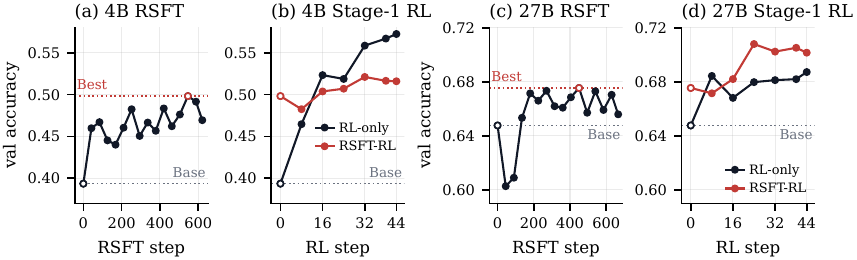}
\caption{Stage-1 accuracy on the 125-question held-out validation set for \qwenS and \qwenL. Filled circles show evaluated checkpoints, while open circles at step 0 mark each route's initial policy. RSFT and RL-only start from the base policy, whereas RSFT-RL starts from the selected RSFT checkpoint. Gray dotted lines show base-policy accuracy. In the RSFT panels, the red open circle and dotted line mark the best checkpoint.}
\label{fig:stage1-val}
\end{figure}

\paragraph{Evaluation protocol.}
The held-out validation set contains 125 questions. We sample four rollouts per question at each checkpoint, giving 500 rollouts per checkpoint. Figure~\ref{fig:stage1-val} shows the checkpoint trajectories, and Table~\ref{tab:stage1-acc} breaks down the selected results by answer type. These questions are used only for Stage-1 checkpoint selection and are separate from the report-stage test events.

\begin{table}[!htb]
\caption{Stage-1 results on the held-out validation set. Acc. averages the per-answer score in Table~\ref{tab:verifier-ladder}. Acc. reward averages the reward in Equation~\ref{eq:stage1}, including the tool check and turn penalty. Tool is the required-tool access rate. The remaining columns give mean scores by answer type. Bold marks the within-scale best.}
\label{tab:stage1-acc}
\centering
\apptab
\setlength{\tabcolsep}{2.4pt}
\begin{tabular*}{\linewidth}{@{\extracolsep{\fill}}lccccccc@{}}
\toprule
& \multicolumn{3}{c}{\textbf{Verifier}} & \multicolumn{4}{c}{\textbf{Accuracy by answer type}} \\
\cmidrule(lr){2-4}\cmidrule(lr){5-8}
\textbf{Checkpoint} & \textbf{Acc.} & \textbf{Acc. reward} & \textbf{Tool} & \textbf{Num.} & \textbf{Loc.} & \textbf{Time} & \textbf{Anom.} \\
\midrule
\qwenS & 0.394 & 0.394 & 0.996 & 0.334 & 0.400 & 0.392 & 0.522 \\
\quad RSFT (step 546) & 0.498 & 0.496 & 0.988 & 0.467 & 0.521 & 0.457 & 0.595 \\
\quad RL-only (step 44) & \best{0.572} & \best{0.572} & 0.998 & \best{0.549} & \best{0.562} & \best{0.526} & \best{0.709} \\
\quad RSFT-RL (step 32) & 0.521 & 0.521 & \best{1.000} & 0.473 & 0.512 & 0.491 & 0.688 \\
\blocksep
\qwenL & 0.648 & 0.648 & 0.998 & 0.571 & 0.637 & 0.685 & 0.785 \\
\quad RSFT (step 450) & 0.675 & 0.675 & \best{1.000} & 0.587 & 0.667 & 0.694 & 0.865 \\
\quad RL-only (step 44) & 0.687 & 0.687 & \best{1.000} & 0.592 & 0.650 & \best{0.728} & \best{0.901} \\
\quad RSFT-RL (step 24) & \best{0.708} & \best{0.708} & \best{1.000} & \best{0.636} & \best{0.679} & 0.724 & 0.893 \\
\bottomrule
\end{tabular*}
\end{table}

Required-tool access ranges from 0.988 to 1.000, and the accuracy reward differs from verifier accuracy by at most 0.002. Differences between checkpoints therefore come mainly from answer quality. For \qwenS, RL-only improves all four answer types over the base policy, with the largest gains on numerical (+0.215) and anomaly (+0.187) questions. For \qwenL, RSFT-RL outperforms RL-only on numerical and location questions but is slightly worse on time and anomaly questions. Thus, the benefit of RSFT depends on model scale and question type.

\subsection{Representative Stage-1 trajectories}
\label{app:stage1-examples}
Claude Opus 5.5 first read and summarized the complete logged trajectories. After reviewing its summaries, we manually selected one representative trajectory of the \qwenS RL-only step-44 checkpoint for each Stage-1 answer type and chose the turn-level information shown in the four boxes below. We checked every displayed detail against the original traces. Brackets summarize the model's reasoning, code is abridged to the relevant lines, and observations are condensed from the logs. The examples cover a one-call numerical solution, a repaired indexing error, recovery from an unsuccessful location lookup, and climatology-based anomaly detection.

\begin{reportbox}[BoxGreen]{{Stage-1 trajectory, \qwenS, numerical task, one call}}
\footnotesize
\textbf{Question.} What is the minimum surface temperature in Antarctica?\par\noindent
\hfill\textbf{Required tool. Dataset.}\quad \textcolor{BoxRed}{\textbf{Gold. 233.517 K.}}

\smallskip
\textbf{Turn 1.} [Use \texttt{2m\_temperature}, locate Antarctica, apply the mask, and reduce.]
{\scriptsize
\begin{verbatim}
def run(ds, geolocator, forecaster, climatology, tempestextremes, plotting):
    mask = geolocator.get_mask_by_geofeature('continent', 'Antarctica')
    t2m = ds[0]["2m_temperature"]
    value = t2m.where(mask == True).min().values
    print(f"Minimum 2m temperature: {value} K")
    return {"min_antarctic_temperature": value}
\end{verbatim}
}
\textit{Observation.} Minimum 2m temperature is 233.4048614501953 K.

\smallskip
\textbf{Turn 2.} [Return one number in Kelvin.] \quad \texttt{<solution>233.40486</solution>}

\smallskip
\textit{Verifier.} The relative error is $4.8\times10^{-4}$, so $s=1$. The required-tool condition is satisfied. The rollout uses two turns and one code call with no execution error.
\end{reportbox}

\begin{reportbox}[BoxGreen]{{Stage-1 trajectory, \qwenS, location task, error repaired}}
\footnotesize
\textbf{Question.} Which continent experienced the lowest average specific humidity at 400 hPa?\par\noindent
\hfill\textbf{Required tool. Dataset.}\quad \textcolor{BoxRed}{\textbf{Gold. Antarctica.}}

\smallskip
\textbf{Turn 1.} [Find the 400 hPa level, mask every continent with the geolocator, and average.]
{\scriptsize
\begin{verbatim}
def run(ds, geolocator, forecaster, climatology, tempestextremes, plotting):
    sh = ds[0]['specific_humidity']
    print("Level coordinates:", sh.level.values)
    level_400_idx = np.where(sh['level'].isel(index=0) == 400)[0][0]
    ...
\end{verbatim}
}
\textit{Observation.} The level coordinates are [50 100 150 200 250 300 400 500 600 700 850 925 1000]. The \texttt{np.where} line then raises \texttt{ValueError: Dimensions \{\textquotesingle index\textquotesingle\} do not exist. Expected one or more of (\textquotesingle level\textquotesingle,)}.

\smallskip
\textbf{Turn 2.} [The level coordinate has no time dimension, so index it directly.]
{\scriptsize
\begin{verbatim}
def run(ds, geolocator, forecaster, climatology, tempestextremes, plotting):
    sh = ds[0]['specific_humidity']
    sh_400 = sh.isel(level=int(np.where(sh.level == 400)[0][0])).isel(index=0)
    results = {}
    for continent in geolocator.get_all_locations('continent'):
        mask = geolocator.get_mask_by_geofeature('continent', continent)
        results[continent] = float(sh_400.where(mask).mean())
    lowest = min(results, key=results.get)
    print("Lowest average:", lowest, "with", results[lowest])
    return {"min_continent": lowest, "min_value": results[lowest]}
\end{verbatim}
}
\textit{Observation.} The lowest average is Antarctica with 0.000112. The other continental means range from 0.000114 in Europe to 0.001050 in South America.

\smallskip
\textbf{Turn 3.} [Return the name and the value in the required JSON form.] \quad \texttt{<solution>\{"name": "Antarctica", "value": 0.000112\}</solution>}

\smallskip
\textit{Verifier.} The normalized name matches the gold name, so $s=1$. The required-tool condition is satisfied. The rollout uses three turns and two code calls, correcting one execution error.
\end{reportbox}

\begin{reportbox}[BoxGreen]{{Stage-1 trajectory, \qwenS, time task, lookup recovered}}
\footnotesize
\textbf{Question.} How many hours from the initial timeframe did Ostrobothnia, Finland experience its highest 10-meter V component of wind?\par\noindent
\hfill\textbf{Required tool. Dataset.}\quad \textcolor{BoxRed}{\textbf{Gold. 132 h.}}

\smallskip
\textbf{Turn 1.} [Identify Finland, list its sublocations, and query the expected coordinates of Ostrobothnia.]
{\scriptsize
\begin{verbatim}
def run(ds, geolocator, forecaster, climatology,
        tempestextremes, plotting):
    finland = geolocator.get_geofeature_by_point(
        25.0, 62.0, "country")
    states = geolocator.get_sublocations("country", finland)
    name = geolocator.get_geofeature_by_point(
        22.0, 64.0, "state")
    print("States in Finland:", states)
    print("Ostrobothnia state:", name)
    return {"finland": finland, "states": states}
\end{verbatim}
}
\textit{Observation.} The direct coordinate lookup returns \texttt{None}. The sublocation list contains the registered name \texttt{Ostrobothnia, Finland}.

\smallskip
\textbf{Turn 2.} [Use the registered name and check that its mask aligns with the wind field.]
{\scriptsize
\begin{verbatim}
def run(ds, geolocator, forecaster, climatology,
        tempestextremes, plotting):
    name = "Ostrobothnia, Finland"
    mask = geolocator.get_mask_by_geofeature("state", name)
    wind = ds[0]["10m_v_component_of_wind"]
    print("Mask shape:", mask.shape)
    print("Wind dimensions:", wind.dims)
    print("Number of time steps:", wind.sizes["index"])
    return {"name": name, "mask_shape": mask.shape}
\end{verbatim}
}
\textit{Observation.} The mask has shape $(360,181)$, matching the longitude and latitude axes of the wind field. The episode contains 28 time steps.

\smallskip
\textbf{Turn 3.} [Find the spatial maximum within the mask at every time step and retain the largest one.]
{\scriptsize
\begin{verbatim}
mask = geolocator.get_mask_by_geofeature(
    "state", "Ostrobothnia, Finland")
wind = ds[0]["10m_v_component_of_wind"]
max_v, max_idx = -1e99, -1
for idx in range(wind.sizes["index"]):
    masked = wind.isel(index=idx) * mask
    value = float(masked.max())
    print(idx, value)
    if value > max_v:
        max_v, max_idx = value, idx
return {"max_v": max_v, "time_index": max_idx,
        "hours": max_idx * 6}
\end{verbatim}
}
\textit{Observation.} The per-time output is long enough that the final lines are truncated, although the execution succeeds.

\smallskip
\textbf{Turn 4.} [Repeat the same reduction without printing every time step.]
{\scriptsize
\begin{verbatim}
mask = geolocator.get_mask_by_geofeature(
    "state", "Ostrobothnia, Finland")
wind = ds[0]["10m_v_component_of_wind"]
max_v, max_idx = -1e99, -1
for idx in range(wind.sizes["index"]):
    value = float((wind.isel(index=idx) * mask).max())
    if value > max_v:
        max_v, max_idx = value, idx
return {"max_v": max_v, "time_index": max_idx,
        "hours": max_idx * 6}
\end{verbatim}
}
\textit{Observation.} The maximum is 8.7056~m\,s$^{-1}$ at index 22, which is 132 hours from the initial time.

\smallskip
\textbf{Turn 5.} [Return the requested hour.] \quad \texttt{<solution>132</solution>}

\smallskip
\textit{Verifier.} The time matches exactly, so $s=1$. The required-tool condition is satisfied. The rollout uses five turns and four code calls with no execution error.
\end{reportbox}

\begin{reportbox}[BoxGreen]{{Stage-1 trajectory, \qwenS, anomaly task, one call}}
\footnotesize
\textbf{Question.} Which water bodies contain time-mean sea-level pressure at least three standard deviations below the all-time climatology?\par\noindent
\hfill\textbf{Required tool. Climatology.}\quad \textcolor{BoxRed}{\textbf{Gold. Philippine Sea.}}

\smallskip
\textbf{Turn 1.} [Standardize the time-mean field against the all-time climatology, then intersect cells below $-3\sigma$ with each water-body mask.]
{\scriptsize
\begin{verbatim}
def run(ds, geolocator, forecaster, climatology,
        tempestextremes, plotting):
    field = ds[0]["mean_sea_level_pressure"].mean(dim="index")
    mu = climatology.get_climatology(
        "mean_sea_level_pressure", "all_time", "mean")
    sigma = climatology.get_climatology(
        "mean_sea_level_pressure", "all_time", "std")
    below = ((field.values - mu.values) / sigma.values) < -3
    answer = []
    for name in geolocator.get_all_locations("water_body"):
        mask = geolocator.get_mask_by_geofeature("water_body", name)
        if (below & mask).any():
            answer.append(name)
    return answer
\end{verbatim}
}
\textit{Observation.} The call returns \texttt{["Philippine Sea"]}.

\smallskip
\textbf{Turn 2.} [Return the anomaly set as JSON.] \quad \texttt{<solution>["Philippine Sea"]</solution>}

\smallskip
\textit{Verifier.} The set F1 is 1. The required-tool condition is satisfied. The rollout uses two turns and one code call with no execution error.
\end{reportbox}

\FloatBarrier

\section{Method and evaluation details}
\label{app:details}

\subsection{Report-stage reward and training constants}
\label{app:reward-constants}
\begin{table}[!t]
\caption{Training and evaluation settings. Shared RL settings apply to Stage-1 RL and report RL, and report-stage settings apply to report RL under both routes.}
\label{tab:reward-constants}
\centering
\apptab
\begin{tabular*}{\linewidth}{@{\extracolsep{\fill}}L{0.29\linewidth}L{0.24\linewidth}L{0.43\linewidth}@{}}
\toprule
\textbf{Setting} & \textbf{Value} & \textbf{Role} \\
\midrule
\catrow{3}{Rejection sampling}
Base-policy rollouts per question & 8 & candidate trajectories for RSFT \\
Sampling & temperature 0.8 & diversify successful trajectories \\
Retention cap & 2 per question & limit repeated successes \\
\midrule
\catrow{3}{Shared RL optimization}
Sampling & temperature 0.6, top-$p$ 1.0 & rollout sampling \\
Rollouts per prompt & 4 & group for the centered advantage \\
Prompts per step & 16 & 64 rollouts per optimizer step \\
Maximum turns & 13 & turn 13 is the forced final turn \\
KL coefficient & 0 & no reference-policy term \\
LoRA rank / alpha & 64 / 64 & language linear layers with frozen vision tower \\
Learning rate, Stage 1 & $10^{-5}$ & LoRA adapter updates \\
Learning rate, report RL & $10^{-5}$ (4B), $3\times10^{-5}$ (27B) & LoRA adapter updates \\
Loss & CISPO & clipped importance ratio, sequence mean \\
CISPO offsets $\epsilon_{\mathrm{low}} / \epsilon_{\mathrm{high}}$ & 1.0 / 4.0 & clipped ratio interval $[0,5]$ \\
\midrule
\catrow{3}{Report-stage reward}
Quality weighting & online, every step & inverse within-group standard deviation \\
Variance smoothing $\beta$ & 0.9 & moving average across steps \\
Weight bounds & $[0.2,\,0.8]$ & keep both scores in the reward \\
Minimum scored groups & 8 & fewer groups keep the previous weight \\
Terminal penalty $c$ & 0.05 & ungrounded or unusable final report \\
Format penalty $p_{\text{fmt}}$ & 0.02 & report recovered from an untagged forced reply \\
Minimum usable report & 40 characters & shorter training reports receive $-c$ without judging \\
\midrule
\catrow{3}{Report-stage episode budgets}
Observation cap & 5{,}000 characters & printed output shown per turn \\
Evidence-summary cap & 2{,}600 characters & server summary appended per turn \\
Prompt token budget & 38{,}000 & forces the final turn when exceeded \\
Episode token budget & 46{,}000 & forces the final turn when exceeded \\
Generation cap & 10{,}000 tokens & one policy action \\
Images per turn / episode & 2 / 8 & long side resized to 768 pixels \\
\midrule
\catrow{3}{Report test evaluation}
Rollouts per event & 1 & one report per system \\
Evaluation cadence & every 32 steps; step 163 & each checkpoint scored by the three judges \\
Sampling & temperature 0.6, top-$p$ 1.0 & same as training for the Qwen policies \\
\bottomrule
\end{tabular*}
\end{table}

The quality reward follows Equation~\ref{eq:quality}, and Table~\ref{tab:reward-constants} lists the reward and training constants. \Direct and \Curr use the same update rule, smoothing, and bounds. The realized weights depend on the rollout scores in each run.
At optimization step $t$, let $G_t$ be the number of prompt groups whose $K$ rollouts all satisfied the grounding requirement and were scored. For component $m\in\{\mathrm{r},\mathrm{f}\}$ (rubric or Discussion F1), the pooled within-group variance is
\begin{equation}
\hat\sigma^2_{m,t} = \frac{1}{G_t(K-1)}\sum_{j=1}^{G_t}\sum_{k=1}^{K}\big(s_{m,jk}-\bar s_{m,j}\big)^2,
\label{eq:pooled-var}
\end{equation}
where $s_{m,jk}$ is the component-$m$ score of rollout $k$ in group $j$ and $\bar s_{m,j}$ its prompt-group mean. An exponential moving average $v_{m,t}=\beta v_{m,t-1}+(1-\beta)\hat\sigma^2_{m,t}$ smooths the estimates.

\noindent With $\tilde\sigma_{m,t}=\max(\sqrt{v_{m,t}},10^{-6})$, the rubric weight is
\begin{equation}
w_t=\operatorname{clip}_{[w_{\min},\,1-w_{\min}]}\!\left(\frac{\tilde\sigma_{\mathrm{f},t}}{\tilde\sigma_{\mathrm{r},t}+\tilde\sigma_{\mathrm{f},t}}\right).
\label{eq:weight}
\end{equation}
Because advantages are centered within each prompt group, the estimate uses within-group variation rather than differences in difficulty across prompts. The inverse-spread rule gives the less variable component a larger coefficient, preventing the more variable score from dominating group-relative advantages. We use $\beta=0.9$ to dampen changes caused by an atypical batch and $w_{\min}=0.2$ to keep both quality signals active. A step updates the weights only when $G_t\geq 8$; otherwise, it retains the previous weights.
Before clipping, the weighted spreads satisfy $w_t\tilde\sigma_{\mathrm{r},t}=(1-w_t)\tilde\sigma_{\mathrm{f},t}$. We apply the blend before computing group-centered advantages. Groups with a grounding or infrastructure failure are excluded so that fixed penalties and missing scores do not distort the estimated quality spread. The weighting affects only the training reward; the raw rubric and Discussion F1 scores are unchanged and reported separately.
Rollouts without a usable, grounded report receive the terminal penalty $-c$. We subtract the format penalty $p_{\text{fmt}}=0.02$ only when the report must be recovered from an untagged forced reply.
\begin{figure}[t]
\centering
\includegraphics[width=\linewidth]{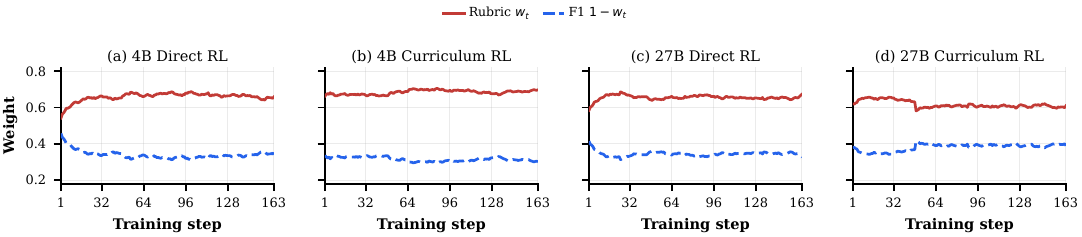}
\caption{Online rubric and Discussion F1 weights during report RL for both routes and model scales. Each run extends through step 163, and the two weights sum to one.}
\label{fig:online-weight}
\end{figure}

Figure~\ref{fig:online-weight} shows the realized weights. For \qwenS, the rubric weight ranges from 0.54 to 0.69 under \Direct and from 0.66 to 0.71 under \Curr. The corresponding ranges for \qwenL are 0.58 to 0.69 and 0.58 to 0.66. Discussion F1 has greater within-group variation and therefore receives less weight throughout each run.

The 163 steps make one pass over the 2,608 report-training events and produce 10,432 rollouts. \Direct and \Curr share the same rollout budget, optimizer, weighting rule, and evaluation schedule. They differ only in their initial policy. Failed executions carry no retry penalty, so the model can repair its code within the turn and resource limits.

\subsection{Report grounding requirement}
\label{app:agent-contract}
At each turn, the model returns either Python code in an \texttt{<execute>} block or a report in a \texttt{<solution>} block. A response containing both tags is rejected without executing the code or ending the episode. A response containing neither tag receives a corrective observation. \Direct and \Curr use the same prompt, which provides the tool documentation, data layout, units, date context, budgets, grounding requirement (called the tool-evidence contract in the prompt), and report style. Both routes allow 12 working turns followed by a forced final turn. Appendix~\ref{app:agent-prompt} reproduces the prompt and environment messages.

The grounding state is maintained by the execution server and reset at the start of every episode. It contains three cumulative Boolean conditions. Dataset access passes after the dataset tool returns at least one valid result. Forecast access passes after the oracle forecaster returns valid fields at a requested lead. Evidence delivery passes after the server appends at least one verified finding to the model's observation. These conditions may be satisfied on different turns, and the grounding flag is their conjunction:
\begin{equation}
g(\tau)=\mathbf{1}[\text{dataset access}]\,
          \mathbf{1}[\text{forecast access}]\,
          \mathbf{1}[\text{evidence delivered}].
\label{eq:grounding-implementation}
\end{equation}
The server updates these conditions from execution metadata rather than from the model's code, printed text, or use of an \texttt{<execute>} tag. Tool access requires a server-recorded successful return. Evidence delivery requires a server-generated finding that is actually included in the observation after output truncation.

Before the forced final turn, a \texttt{<solution>} with $g(\tau)=0$ is rejected, consumes the turn, and receives a message requesting the missing evidence. The episode then continues. At the cutoff, the environment requests a final response regardless of the grounding state. If $g(\tau)=0$ or the final response is not a usable report of at least 40 characters, the rollout receives the fixed reward $-c=-0.05$. Judge scores for a usable but ungrounded report are recorded only for monitoring, and unusable reports are not judged. During offline evaluation, a final report with $g(\tau)=0$ is likewise treated as a failure and replaced by the no-report sentinel before judging. Unambiguous server or judge failures are handled separately by excluding the affected prompt group from the update.

\subsection{Representative report-stage trajectory}
\label{app:report-trajectory}
The trajectory below produced the \qwenL \Curr report in Figure~\ref{fig:reportviz}. We selected the displayed turns and information using the same procedure as for the Stage-1 trajectories in Appendix~\ref{app:stage1-examples}. The box shows the two evidence-acquisition calls, summarizes the remaining analysis, and reproduces the opening paragraph of the final report.

\begin{reportbox}[BoxBlue]{{Report-stage trajectory, \qwenL \Curr}}
\footnotesize
\textbf{Request.} Produce an approximately 300-word operational CONUS report covering systems and movement, temperature, precipitation and hazards, pressure evolution, fronts, and the jet stream.

\smallskip
\textbf{Turn 1. Analysis acquisition.} [Read the current pressure and 500 hPa fields over CONUS.]
{\scriptsize
\begin{verbatim}
def run(ds, geolocator, forecaster, climatology,
        tempestextremes, plotting):
    d = ds[0]
    mslp = d["mean_sea_level_pressure"].isel(index=0) / 100
    us_mslp = mslp.sel(latitude=slice(24, 50),
                        longitude=slice(235, 293))
    pmin = us_mslp.isel(
        us_mslp.argmin(dim=["longitude", "latitude"]))
    z500 = d["geopotential"].sel(level=500).isel(index=0) / 9.80665
    return {"mslp_min": float(pmin),
            "z500_min": float(z500.sel(latitude=slice(24, 50),
              longitude=slice(235, 293)).min())}
\end{verbatim}
}
\textit{Observation.} Current MSLP minimum 1008.33 hPa near $50^\circ$N, $110^\circ$W and 500 hPa minimum 5429.72 m. Server-delivered evidence additionally records 70.44 mm maximum 12-hour precipitation in the southern Plains, a 65.49 m/s 250 hPa jet maximum in the northern Plains, and a 5891.5 m 500 hPa maximum in the Southeast. Dataset access and evidence delivery now pass; forecast access does not yet pass.

\smallskip
\textbf{Turn 2.} [Plot current MSLP and 500 hPa height to inspect the trough-ridge structure.]

\smallskip
\textbf{Turn 3. Forecast acquisition.} [Generate 6--72 h oracle fields and detect extratropical tracks.]
{\scriptsize
\begin{verbatim}
def run(ds, geolocator, forecaster, climatology,
        tempestextremes, plotting):
    context = ds[0].isel(index=[0, 2])
    preds = [forecaster.forecast(context, horizon=h)
             for h in range(1, 13)]
    seq = xr.concat(preds, dim="index")
    tracks = tempestextremes.detect_with_profile(
        seq, mode="extratropical", strictness="medium",
        region={"minlat": 24, "maxlat": 52,
                "minlon": -125, "maxlon": -70}, top_k=4)
    return {"n_tracks": len(tracks)}
\end{verbatim}
}
\textit{Observation.} Twelve forecast leads and two tracks returned. Verified evidence includes a t+36\,h MSLP minimum of 1005.72 hPa and 500 hPa minimum of 5395.6 m in the northern Plains, plus 50.13 mm maximum 12-hour precipitation in the southern Plains. Dataset access, forecast access, and evidence delivery now all pass.

\smallskip
\textbf{Turns 4--6.} [Inspect track points, 850 hPa temperature, precipitation, and the t+36\,h 500 hPa field.] All three executions succeed. The agent interprets an eastward-moving northern trough and front, colder air behind it, and a persistent heavy-rain signal over the southern Plains.

\smallskip
\textbf{Turn 7. Final report.} [Synthesize the verified analysis and forecast evidence into continuous prose.]

\smallskip
\textit{Opening paragraph.} A longwave 500 mb trough currently sits over the northern Plains and Canadian prairies, with the upper-level jet core (250 mb winds near 65 m/s) overhead and a surface low centered near the northern Plains-Canadian border around 1008 hPa. A cold front extends south from that low across the central Plains, marking a sharp 850 mb thermal gradient between sub-freezing air over the northern Plains and warm air (8--12~°C at 850 mb) over the South. A secondary low over the southern Plains, near the Oklahoma-Kansas-Texas Panhandle area, is tapping warm, moist Gulf air and is the main source of heavy rain, with 12-hour accumulations already reaching 50--70 mm and 6-hour totals near 45 mm.

\smallskip
The complete report appears with claim labels in Figure~\ref{fig:reportviz} and Appendix~\ref{app:judgeviz}. The episode ends after seven turns with six successful executions, no execution errors, a satisfied grounding requirement, and a tagged final report.
\end{reportbox}

\subsection{Discussion F1 scoring}
\label{app:entailment}
Discussion F1 uses the claim extraction prompt and the claim evaluation prompt of Appendix~\ref{app:entailment-prompts}. The judge is queried separately for each claim at temperature zero.
As defined in Section~\ref{sec:reward}, NEUTRAL generated claims are excluded from the precision denominator. Precision and F1 are zero when no generated claim is SUPPORTED or REFUTED, including when no claims are extracted from the generated report. A reference claim that the report neither states nor contradicts counts as a recall miss.
For offline evaluation, reports that are ungrounded ($g(\tau)=0$) or shorter than 40 characters are replaced by the no-report sentinel before claim extraction. The sentinel is processed through extraction and claim labeling in both directions, and the score follows the same precision and recall formulas. During training, unusable reports bypass both judges and receive the terminal penalty $-c$.

\subsection{Rubric construction and scoring}
\label{app:rubric}

\paragraph{Construction.}
Table~\ref{tab:rubric} lists the six criteria in the frozen rubric. We construct
the rubric from human-written references in the report-training
split. We sample 100 references for positive examples and another
100 for negative examples. GPT-5.2 creates each positive by faithfully
paraphrasing its reference. It creates each negative by removing weather
claims, adding claims from an unrelated training reference, or applying both
changes. Light, medium, and heavy corruptions alter one, two to three, and four
to six claims, respectively. We then paraphrase the negatives in the same way as the
positives to reduce stylistic cues. We also sample 25 positive and 25 negative
examples from disjoint tasks for final rubric evaluation. All
source reports come from the report-training split, but these examples are
used only to construct the rubric and never to train the policy.

\begin{table}[!htb]
\caption{The six meteorological rubric criteria supplied to the judge. Anchored integer scores are normalized to $[0,1]$.}
\label{tab:rubric}
\centering
\apptab
\begin{tabular*}{\linewidth}{@{\extracolsep{\fill}}L{0.27\linewidth}L{0.68\linewidth}@{}}
\toprule
\textbf{Criterion} & \textbf{Definition} \\
\midrule
Setup Fidelity \& Key Feature Retention (Additions AND Omissions) & Accurately represents the specific synoptic/mesoscale features and stated hazards/corridors in the setup, without introducing unsupported details and without dropping any primary stated feature/hazard that changes the forecast story. \\
\rowsep
Spatiotemporal Specificity (Progression When Present; Persistence When Static) & Communicates where and when impacts occur with specificity appropriate to the setup: describes progression tied to moving features when the setup is progressive, or explicitly states persistence/limited movement when the setup is quasi-stationary. \\
\rowsep
Impact/Hazard Extraction \& Prioritization (Reference-Weighted) & Identifies the setup's main hazards and impact corridors (and any stated relative maxima/minima) and ranks them appropriately, using practical, decision-relevant framing without adding new hazards. \\
\rowsep
Mechanism-to-Weather Reasoning (Sufficient, Setup-Consistent) & Uses meteorologically correct, setup-consistent causal links (forcing, moisture transport, instability, terrain/onshore/upslope, deformation/lake-effect) to explain why impacts occur where/when described. Depth is rewarded but not required for a high score if the forecast is a faithful paraphrase. \\
\rowsep
Thermal/Airmass \& Precip-Type Plausibility (No Out-of-Pattern P-Type) & Maintains a coherent temperature/airmass narrative consistent with the setup and seasonally plausible precipitation type statements (snow/ice/convective vs stratiform), avoiding thermodynamic regime shifts not supported by the setup. \\
\rowsep
Constraint-Respecting Detail \& Calibration (No Invented Timing/Numbers/Pattern Labels) & Uses specificity only when supported: avoids invented cutoffs (``ends by Saturday evening''), ungrounded pattern labels (``split flow'') or pseudo-precision, and matches the setup's confidence level (possible/expected/likely). \\
\bottomrule
\end{tabular*}
\end{table}

GPT-5.2 writes and scores the candidate rubrics with medium reasoning effort.
The first round receives five positive and five negative seed examples that
are separate from the refinement subset. We score each candidate on a fixed
refinement subset of 16 positive and 16 negative examples. In later rounds, GPT-5.2 sees
the previous rubric, both score distributions, and the six examples with the
most problematic scores. It then revises the criteria and anchors. Refinement
runs for at most four rounds and stops early once the ROC AUC for separating
positives from negatives is at least 0.9 and its round-to-round gain is below 0.05. We select the candidate with the highest AUC and use the mean score
gap to break ties. The selected rubric comes from round 3. We evaluate it once
on the 25 positive and 25 negative held-out examples, review each criterion, and freeze it before
policy training. Examples from the report test set and outputs from policy
checkpoints are never used during rubric construction.

\paragraph{Scoring protocol.}
The scoring prompt contains the frozen rubric, task request, reference
report, and generated report. At temperature zero, one judge call returns an
integer score and a brief explanation for each of the six criteria, followed
by an overall assessment. The rubric anchors scores at 1, 3, 5, 8, and 10 but
allows intermediate integers. We round returned values to integers and
clip them to $[1,10]$. A missing criterion receives the minimum score. The raw
rubric average is the mean of the six scores. We report the normalized value
$(\text{average}-1)/9$.

For offline evaluation, an ungrounded report or one shorter than 40 characters is
replaced by the no-report sentinel and passed through the same judge. During training, only
grounded, usable reports receive judge-based rewards. Ungrounded or unusable
reports receive the terminal penalty $-c$. Each reported score for a system-judge pair
averages three complete passes over the same 197 frozen reports.
Appendix~\ref{app:agent-contract} details the penalty and the monitoring-only scores.

\paragraph{Complete scoring anchors.}\label{app:rubric-anchors}
Figure~\ref{fig:rubric-anchors} gives the complete scoring anchors for the six rubric criteria.
\newcommand{\rubricanchorpanel}[2]{%
\begin{minipage}[t]{0.485\linewidth}
\vspace{0pt}
\begin{tcolorbox}[colback=white,colframe=black!35,colbacktitle=black!10,
coltitle=black,fonttitle=\fontsize{7}{7.7}\selectfont\bfseries\raggedright,
fontupper=\fontsize{7}{7.7}\selectfont,title={#1},
boxrule=0.4pt,arc=1mm,left=3pt,right=3pt,top=1pt,bottom=1pt,
before skip=0pt,after skip=0pt]
\setlength{\tabcolsep}{2pt}
\renewcommand{\arraystretch}{1.0}
\begin{tabular}{@{}rL{0.88\linewidth}@{}}
#2
\end{tabular}
\end{tcolorbox}
\end{minipage}}

\begin{figure}[!t]
\centering
\begingroup
\renewcommand{\rowsep}{\specialrule{0.25pt}{0.4pt}{0.4pt}}
\rubricanchorpanel{Setup Fidelity \& Key Feature Retention (Additions AND Omissions)}{
1 & Multiple major errors: invents or contradicts key systems/hazards OR omits multiple primary hazards/corridors explicitly in the setup. \\
\rowsep
3 & One major fault: a single invented/contradictory feature that materially changes the story (e.g., added cutoff/dry slot/Arctic front) OR omission of a primary stated hazard/corridor (e.g., ignores a named snowstorm area). \\
\rowsep
5 & Broadly consistent, but includes noticeable unsupported embellishments or misses an important secondary detail (still preserves the main story). \\
\rowsep
8 & Faithfully captures all primary features/hazards/corridors; only small, non-impactful deviations or compressions. \\
\rowsep
10 & Near-verbatim fidelity: all primary features/hazards and their described relationships are retained; no unsupported additions. \\
}\hfill
\rubricanchorpanel{Spatiotemporal Specificity (Progression When Present; Persistence When Static)}{
1 & Vague in both time and place; no usable windows or corridors. \\
\rowsep
3 & Some geography or some timing, but not both; progression/persistence is unclear or misleading. \\
\rowsep
5 & Provides identifiable corridors and broad windows (e.g., through Thu/Fri, weekend), but progression/persistence linkage to the setup is incomplete. \\
\rowsep
8 & Clear corridors and windows; correctly describes either (a) feature-tied progression or (b) persistent/unchanged pattern when that is the setup. \\
\rowsep
10 & Consistently precise windows (day/night or 24--48h) and corridor evolution tied to named features, without inventing timing details. \\
}\par\vspace{5pt}
\rubricanchorpanel{Impact/Hazard Extraction \& Prioritization (Reference-Weighted)}{
1 & Misses or misidentifies the primary hazards, or prioritizes a non-existent hazard as the main threat. \\
\rowsep
3 & Mentions hazards but misses a key high-impact corridor/maxima stated in the setup OR substantially downplays it into generic wording. \\
\rowsep
5 & Captures the main hazard types and broad areas, but prioritization and `where it's worst' localization are weak or generic. \\
\rowsep
8 & Clearly highlights primary hazard corridor(s) with timing and relative emphasis consistent with the setup; separates main vs secondary impacts. \\
\rowsep
10 & Decision-ready: includes the setup's key qualifiers where given (e.g., explicit `few feet'/`south of X line' style maxima, `possible vs expected') and avoids adding unstated threats. \\
}\hfill
\rubricanchorpanel{Mechanism-to-Weather Reasoning (Sufficient, Setup-Consistent)}{
1 & Mechanisms are largely incorrect or contradict the setup (e.g., wrong forcing for stated precip type/region). \\
\rowsep
3 & Mostly outcome listing with weak/incorrect causal links, or mechanisms are used to justify details not present in the setup. \\
\rowsep
5 & Basic correct links (e.g., ridge$\rightarrow$heat, onshore/upslope$\rightarrow$mountain precip, boundary$\rightarrow$convection) but limited differentiation of intensity/coverage. \\
\rowsep
8 & Multiple correct mechanisms tied to specific corridors/timing, explaining distribution and intensity differences without overreach. \\
\rowsep
10 & Integrated, physically grounded explanation (moisture/forcing/thermodynamics/terrain) that clearly supports the stated impacts and evolution. \\
}\par\vspace{5pt}
\rubricanchorpanel{Thermal/Airmass \& Precip-Type Plausibility (No Out-of-Pattern P-Type)}{
1 & Thermal/ptype narrative is implausible or contradictory (e.g., widespread snow/ice or major cold outbreak implied without support). \\
\rowsep
3 & Introduces a notable but unsupported thermodynamic/ptype implication (e.g., adds a cold-air push driving wintry precip outside stated areas) OR omits a key stated p-type transition (e.g., snow/wintry mix explicitly mentioned). \\
\rowsep
5 & Mostly plausible; limited temperature evolution detail, but no major thermodynamic errors. \\
\rowsep
8 & Coherent regional thermal contrasts/trends consistent with the setup; precip type aligned with elevation/airmass where relevant. \\
\rowsep
10 & Strong, setup-faithful airmass evolution with clear implications for sensible weather and precip type where the setup supports it. \\
}\hfill
\rubricanchorpanel{Constraint-Respecting Detail \& Calibration (No Invented Timing/Numbers/Pattern Labels)}{
1 & Frequent unjustified specifics (exact timing, magnitudes, or dynamics) or overconfident tone that is not supported by the setup. \\
\rowsep
3 & At least one impactful unsupported embellishment (e.g., a specific weakening time, moisture cutoff, added structural diagnosis) that meaningfully changes interpretation or confidence. \\
\rowsep
5 & Mostly calibrated; occasional minor over-specific or extra-structural wording, but not dominant. \\
\rowsep
8 & Well-calibrated language; any numbers/timing/structure used are clearly supported by the setup; no fake precision. \\
\rowsep
10 & Exemplary discipline: mirrors the setup's certainty, preserves stated qualifiers, and adds no extra structure beyond what is warranted. \\
}
\endgroup
\caption{Complete scoring anchors for the six meteorological rubric criteria. The judge receives the anchor descriptions shown for levels 1, 3, 5, 8, and 10. Intermediate integer scores are allowed.}
\label{fig:rubric-anchors}
\end{figure}

High scores require faithful coverage of the main features, hazards, locations, timing, and uncertainty. Additional mechanisms or numerical details help only when the setup supports them. The rubric therefore rewards accurate synthesis rather than detail alone.

\subsection{Training details}
\label{app:infra}
\paragraph{Policy objective.}
For each prompt, the sampler draws $K$ rollouts, each a single sequence containing the full interaction history. The policy is updated with the sequence-mean CISPO loss of Equation~\ref{eq:cispo} in Section~\ref{sec:optim}. LoRA adapters are applied to the language layers, the vision tower is frozen, and the KL coefficient is zero. Observations, evidence summaries, and forced-turn messages are environment tokens and are excluded from the loss.

\paragraph{Failure handling.}
An unambiguous infrastructure or judge failure excludes its four-rollout prompt group from the update. Code errors and grounding failures caused by the policy are retained as training outcomes.

\paragraph{Hardware.}
Each training run uses two NVIDIA H200 GPUs. For the 4B model, rollout inference and policy optimization run on separate devices; for the 27B model, a two-way tensor-parallel inference engine and two policy ranks share both devices. Standalone inference and evaluation use NVIDIA A100 GPUs.

\subsection{Checkpoint selection and evaluation protocol}
\label{app:selection}
We select Stage-1 checkpoints by mean verifier credit on the 125-question held-out validation set. If two checkpoints have the same mean verifier credit, we select the one with the higher mean accuracy reward (Equation~\ref{eq:stage1}) and, if they are still tied, the later training step. Report-stage comparisons use the terminal step-163 policy. We also evaluate steps 32, 64, 96, 128, and 160 but omit step 160 from the curves. Each judge scores the same frozen reports at a given checkpoint. These diagnostics use the 197 test events, but they do not determine any checkpoint or hyperparameter. Appendix~\ref{app:checkpoints} shows the plotted trajectories.

For the tool-use diagnostics in Figure~\ref{fig:effects}, the server records, for each execution request, the set of distinct tool families that returned valid results. Tools/exec. counts these families, while failure rate is the fraction of requests for which this set is empty. We first average each measure within an episode and then across the 197 episodes, so episodes with more requests do not receive extra weight. An episode with no execution request receives a failure rate of one.

\section{Additional results}
\label{app:results}

This appendix reports checkpoint trajectories, judge diagnostics, and the expert audit of claim labels.

\subsection{Checkpoint trajectories}
\label{app:checkpoints}
\begin{figure}[!htbp]
\centering
{\small (a) \qwenS}\\[1pt]
\includegraphics[width=0.78\linewidth,trim=0 10 0 0,clip]{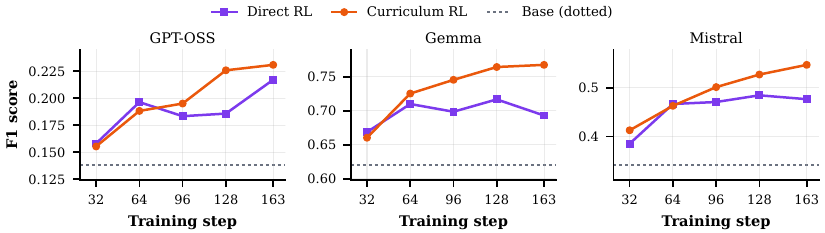}\\
\includegraphics[width=0.78\linewidth,trim=0 0 0 12,clip]{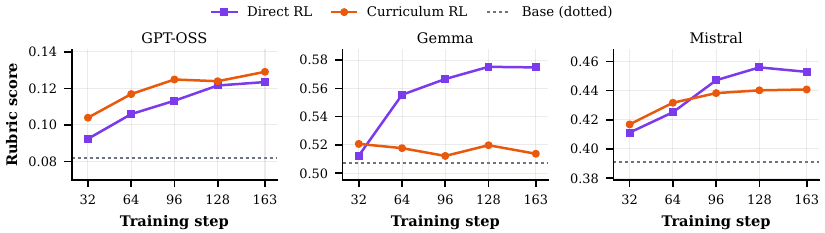}\\[6pt]
{\small (b) \qwenL}\\[1pt]
\includegraphics[width=0.78\linewidth,trim=0 10 0 12,clip]{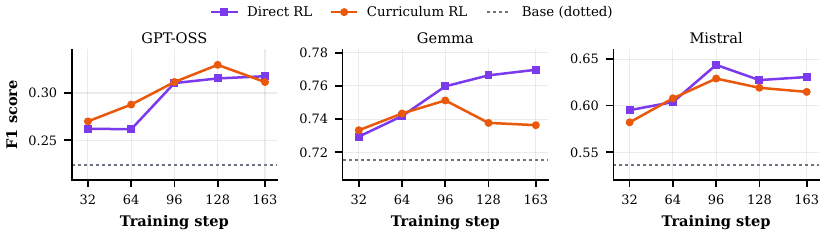}\\
\includegraphics[width=0.78\linewidth,trim=0 0 0 12,clip]{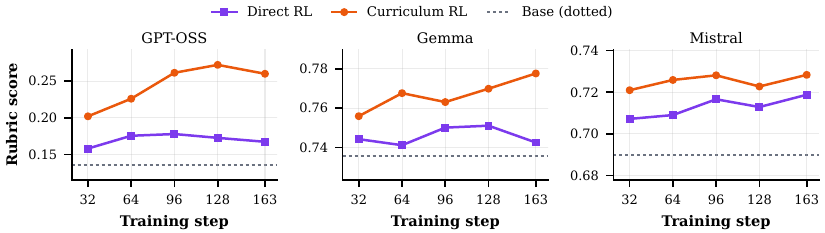}
\caption{Report-RL F1 and rubric-score trajectories for (a) \qwenS and (b) \qwenL across the three evaluation judges. Each point averages three scoring passes on the 197 test events.}
\label{fig:checkpoints}
\end{figure}

Figure~\ref{fig:checkpoints} tracks F1 and rubric score for both routes at every evaluated checkpoint. Every checkpoint exceeds the base policy in F1 under all three judges, so report RL improves claim-level agreement from the first evaluation onward. At 4B, \Curr leads in F1 from step 96 onward, whereas the independent-judge rubric curves generally favor \Direct, matching the terminal results in Table~\ref{tab:main}. At 27B, \Curr has the higher rubric score at every checkpoint under all three judges, while the F1 curves cross during training and \Direct finishes slightly ahead. All main comparisons use the terminal step-163 policies rather than intermediate peaks.

\subsection{Judge diagnostics}
\label{app:judges}
Every judge call uses temperature zero, but batched GPU inference is not fully deterministic, so repeated passes over the same frozen reports can return different claims, labels, and scores. We therefore run three complete scoring passes per judge and report their mean.

The judges disagree more on individual reports than on overall system rankings (Table~\ref{tab:judge-agreement}). \gemma and \mistraljudge agree most on both report-level metrics and the rubric ranking, whereas \oss and \mistraljudge agree most on the F1 ranking.

Table~\ref{tab:judge-labels} explains some of these differences. \mistraljudge extracts the most claims, \gemma assigns the largest supported fractions, and \oss is the most conservative on both report and reference claims. Each judge determines both the claims and their labels, so these choices change the F1 numerator and denominator together. Route comparisons within a judge are therefore more informative than comparisons of raw scores across judges. Appendix~\ref{app:judgeviz} shows the three labeling patterns on one report.

\begin{table}[!htb]
\centering
\apptab
\setlength{\tabcolsep}{2.4pt}
\begin{minipage}[t]{0.48\linewidth}
\caption{Agreement between judges across 15 systems and 2,955 system-event reports. Per-report values are Spearman correlations between three-pass mean scores. Ranking values are Kendall's $\tau$ between system rankings.}
\label{tab:judge-agreement}
\centering
\begin{tabular*}{\linewidth}{@{\extracolsep{\fill}}lcccc@{}}
\toprule
& \multicolumn{2}{c}{\textbf{Per report}} & \multicolumn{2}{c}{\textbf{Ranking}} \\
\cmidrule(lr){2-3}\cmidrule(lr){4-5}
\textbf{Judge pair} & \textbf{Rubric} & \textbf{F1} & \textbf{Rubric} & \textbf{F1} \\
\midrule
\oss{} / \gemma & 0.46 & 0.48 & 0.66 & 0.71 \\
\oss{} / \mistraljudge & 0.49 & 0.61 & 0.77 & 0.85 \\
\gemma{} / \mistraljudge & 0.75 & 0.62 & 0.89 & 0.71 \\
\bottomrule
\end{tabular*}
\end{minipage}\hfill
\begin{minipage}[t]{0.50\linewidth}
\caption{Claim extraction and support rates for the same 15 systems as Table~\ref{tab:judge-agreement}. Values pool all three scoring passes. Per rep. is the mean claim count per report, while Supp. and Cov. are the supported fractions of report and reference claims, respectively.}
\label{tab:judge-labels}
\centering
\begin{tabular*}{\linewidth}{@{\extracolsep{\fill}}lcccc@{}}
\toprule
& \multicolumn{2}{c}{\textbf{Report claims}} & \multicolumn{2}{c}{\textbf{Reference claims}} \\
\cmidrule(lr){2-3}\cmidrule(lr){4-5}
\textbf{Judge} & \textbf{Per rep.} & \textbf{Supp.} & \textbf{Per rep.} & \textbf{Cov.} \\
\midrule
\oss & 11.7 & 0.15 & 12.8 & 0.20 \\
\gemma & 10.3 & 0.58 & 11.4 & 0.68 \\
\mistraljudge & 16.6 & 0.28 & 14.6 & 0.49 \\
\bottomrule
\end{tabular*}
\end{minipage}
\end{table}

\subsection{Expert audit of claim labels}
\label{app:alignment}
All three meteorologists label the same 100 claim-report pairs from five randomly sampled report-generation tasks, with claims extracted by \oss. The reports come from \qwenS, \qwenL, and GPT-5.6-Luna. Each task includes both comparison directions, with generated claims checked against the reference and reference claims against the generated report. The meteorologists see neither the direction nor the report generator.

The meteorologists use the same SUPPORTED, NEUTRAL, and REFUTED definitions as the LLM judges. Accuracy counts exact label matches, while Cohen's $\kappa$ adjusts for chance agreement. Within-1 accuracy treats only direct SUPPORTED-REFUTED disagreements as mismatches. We compute majority labels separately for the three meteorologists and three LLM judges, assigning NEUTRAL to a three-way tie. Table~\ref{tab:expert-claim-agreement} reports pairwise and majority-vote agreement.

\FloatBarrier
\section{Judge claim labels for one report}
\label{app:detail}

\label{app:judgeviz}
Figures~\ref{fig:judgeviz-gemma} and~\ref{fig:judgeviz-mistral} map the \gemma and \mistraljudge claim labels onto the report in Figure~\ref{fig:reportviz}.
\begin{figure}[H]
\centering
\begingroup\microtypesetup{expansion=false}
\setlength{\parindent}{0pt}%
\providecolor{PairSlate}{HTML}{475569}%
\tcbset{pairreport/.style args={#1}{enhanced,nobeforeafter,colback=white,
  colbacktitle=#1,colframe=#1,coltitle=white,
  fonttitle=\scriptsize\bfseries,boxrule=0.55pt,arc=1.2mm,
  left=3.5pt,right=3.5pt,top=1pt,bottom=1pt,
  toptitle=1pt,bottomtitle=1pt,boxsep=0pt}}%
\begin{tcolorbox}[pairreport=PairSlate,title={Curriculum RL report with Gemma-3-27B-IT claim labels}]
\fontsize{8.5}{8.6}\selectfont\setlength{\parskip}{1pt}%
\SclaimBg{1}{A longwave 500 mb trough currently sits over the northern Plains and Canadian prairies, with the upper-level jet core (250 mb winds near 65 m/s) overhead and a surface low centered near the northern Plains-Canadian border around 1008 hPa.} \SclaimBg{2}{A cold front extends south from that low across the central Plains, marking a sharp 850 mb thermal gradient between sub-freezing air over the northern Plains and warm air (8--12~°C at 850 mb) over the South.} \SclaimBg{3}{A secondary low over the southern Plains, near the Oklahoma-Kansas-Texas Panhandle area, is tapping warm, moist Gulf air and is the main source of heavy rain, with 12-hour accumulations already reaching 50--70 mm and 6-hour totals near 45 mm.}

\SclaimBg{4}{Over the next 24--48 hours the upper trough and its surface low will track east across the northern Plains and into the Upper Midwest and Great Lakes, bringing a broad band of showers and storms to the northern Plains, the Dakotas, and the Lakes through Friday into Saturday.} \SclaimBg{5}{Cooler air behind the front will cap daytime highs and allow snow showers along the northern Rockies and higher terrain where 850 mb temperatures remain below freezing.} \SclaimBg{6}{Meanwhile the southern Plains low and front will persist into the weekend, maintaining a risk of scattered to frequent thunderstorms with locally heavy rain, flash-flood potential, and the threat of severe weather given the warm, moist airmass and strong mid-level flow.}

\RclaimBg{1}{The Southeast and much of the South remain under the eastern flank of a 500 mb ridge, where conditions stay warm, stable, and largely dry, with 2 m temperatures in the low to mid 30s~°C and only isolated afternoon activity.} \SclaimBg{7}{The Pacific Northwest and northern Rockies will see intermittent precipitation from the western flank of the trough.} \SclaimBg{8}{The dominant hazard through the period is therefore the Plains frontal system: a moving band of precipitation and cool-air advection across the north, and a persistent, potentially severe and flood-prone storm track across the southern Plains, while the South enjoys warm, ridge-dominated weather.}
\end{tcolorbox}\par
\vspace{1pt}
\begin{tcolorbox}[pairreport=PairSlate,title={Reference report with evidence for SUPPORTED and REFUTED claims}]
\fontsize{8.5}{8.6}\selectfont\setlength{\parskip}{1pt}%
The meteorological discussion and forecast for the continental United States is: \EsentR{1}{Severe thunderstorms and heavy rainfall with the threat of flash flooding are expected for parts of the central and eastern U.S.}

A front stretched across the Ohio and Tennessee Valleys southwestward into the Lower Mississippi Valley and Southern Plains is expected to slowly make its way eastward across the eastern U.S. through the end of the week. \EsentBg{3}{Moisture from the Gulf of Mexico will cause widespread rain and thunderstorms along and ahead of the front.} \EsentBg{6}{Severe weather and heavy rain causing flash flooding are both possible.} \EsentBg{6}{On Friday, the most likely area for severe weather will be in southern Texas.} \EsentBg{6}{There is also a risk of excessive rainfall and flash flooding for much of Texas and into Louisiana and Arkansas.} Isolated strong thunderstorms and heavy rainfall with the possibility of causing flooding are possible into the Ohio Valley and Mid-Atlantic region as well. On Saturday, heavier rainfall totals are expected across the Ohio Valley, with a risk of flash flooding there. \EsentR{1}{The severe weather threat is forecast farther south across the Southeast and Southern Appalachians.}

\EsentBg{2}{The front will divide warmer than average temperatures to the east of it and cooler than average temperatures to the west.} \EsentBg{8}{The Southeast and Tennessee Valley will be 5 to 10 degrees above average and the Mid-Atlantic region 10 to 15 degrees above average for high temperatures on Friday.} A backdoor front across the northern Mid-Atlantic on Friday will keep the Northeast cooler than average. \EsentBg{2}{Behind the main front, temperatures are forecast to be 5 to 15 degrees cooler than average in the Southern/Central Plains and into the Middle Mississippi Valley.} As the front moves eastward on Saturday, the Ohio Valley could see high temperatures 5 to 10 degrees below average.

\EsentBg{6}{Farther west, scattered thunderstorms are forecast to develop on Saturday in the Southern and Central High Plains, which could be severe.} \EsentBg{1, S4}{A cold front coming into the Northern Plains late Friday} \EsentBg{4}{will progress into the Upper Midwest on Saturday, causing showers along it.} \EsentBg{5}{Temperatures are expected to cool below average in the Northern Rockies and Plains with this front,} \EsentBg{7, S5}{which could lead to light snow in those areas on Saturday.} The rest of the West will warm and have above average temperatures for the most part by Saturday, and generally remain dry ahead of a low approaching California by Sunday morning.
\end{tcolorbox}
\caption{Report from Figure~\ref{fig:reportviz} with every \gemma claim labeled SUPPORTED (\textcolor{ClaimSup}{\textbf{S}}), REFUTED (\textcolor{ClaimRef}{\textbf{R}}), or NEUTRAL (\textbf{N}). SUPPORTED and REFUTED claims are shaded green and red, and the reference sentences that support or refute them carry the same tag and shading.}
\label{fig:judgeviz-gemma}
\endgroup
\end{figure}

\begin{figure}[H]
\centering
\begingroup\microtypesetup{expansion=false}
\setlength{\parindent}{0pt}%
\providecolor{PairSlate}{HTML}{475569}%
\tcbset{pairreport/.style args={#1}{enhanced,nobeforeafter,colback=white,
  colbacktitle=#1,colframe=#1,coltitle=white,
  fonttitle=\scriptsize\bfseries,boxrule=0.55pt,arc=1.2mm,
  left=3.5pt,right=3.5pt,top=1pt,bottom=1pt,
  toptitle=1pt,bottomtitle=1pt,boxsep=0pt}}%
\begin{tcolorbox}[pairreport=PairSlate,title={Curriculum RL report with Mistral-Small-3.2-24B claim labels}]
\fontsize{8.5}{8.6}\selectfont\setlength{\parskip}{1pt}%
\NclaimBg{1}{A longwave 500 mb trough currently sits over the northern Plains and Canadian prairies,} \NclaimBg{2}{with the upper-level jet core (250 mb winds near 65 m/s) overhead} \NclaimBg{3}{and a surface low centered near the northern Plains-Canadian border around 1008 hPa.} \NclaimBg{4}{A cold front extends south from that low across the central Plains,} \NclaimBg{5}{marking a sharp 850 mb thermal gradient between sub-freezing air over the northern Plains and warm air (8--12~°C at 850 mb) over the South.} \SclaimBg{1}{A secondary low over the southern Plains, near the Oklahoma-Kansas-Texas Panhandle area, is tapping warm, moist Gulf air and is the main source of heavy rain,} \NclaimBg{6}{with 12-hour accumulations already reaching 50--70 mm and 6-hour totals near 45 mm.}

\NclaimBg{7}{Over the next 24--48 hours the upper trough and its surface low will track east across the northern Plains and into the Upper Midwest and Great Lakes,} \NclaimBg{8}{bringing a broad band of showers and storms to the northern Plains, the Dakotas, and the Lakes through Friday into Saturday.} \SclaimBg{2}{Cooler air behind the front will cap daytime highs and allow snow showers along the northern Rockies and higher terrain where 850 mb temperatures remain below freezing.} \SclaimBg{3}{Meanwhile the southern Plains low and front will persist into the weekend, maintaining a risk of scattered to frequent thunderstorms with locally heavy rain, flash-flood potential, and the threat of severe weather given the warm, moist airmass and strong mid-level flow.}

\RclaimBg{1}{The Southeast and much of the South remain under the eastern flank of a 500 mb ridge, where conditions stay warm, stable, and largely dry,} \RclaimBg{2}{with 2 m temperatures in the low to mid 30s~°C and only isolated afternoon activity.} \NclaimBg{9}{The Pacific Northwest and northern Rockies will see intermittent precipitation from the western flank of the trough.} \SclaimBg{4}{The dominant hazard through the period is therefore the Plains frontal system: a moving band of precipitation and cool-air advection across the north, and a persistent, potentially severe and flood-prone storm track across the southern Plains,} \RclaimBg{3}{while the South enjoys warm, ridge-dominated weather.}
\end{tcolorbox}\par
\vspace{1pt}
\begin{tcolorbox}[pairreport=PairSlate,title={Reference report with evidence for SUPPORTED and REFUTED claims}]
\fontsize{8.5}{8.6}\selectfont\setlength{\parskip}{1pt}%
The meteorological discussion and forecast for the continental United States is: \EsentR{1}{Severe thunderstorms and heavy rainfall with the threat of flash flooding are expected for parts of the central and eastern U.S.}

A front stretched across the Ohio and Tennessee Valleys southwestward into the Lower Mississippi Valley and Southern Plains is expected to slowly make its way eastward across the eastern U.S. through the end of the week. \EsentBg{1}{Moisture from the Gulf of Mexico will cause widespread rain and thunderstorms along and ahead of the front.} \EsentBg{3}{Severe weather and heavy rain causing flash flooding are both possible.} \EsentBg{3}{On Friday, the most likely area for severe weather will be in southern Texas.} \EsentBg{3}{There is also a risk of excessive rainfall and flash flooding for much of Texas and into Louisiana and Arkansas.} Isolated strong thunderstorms and heavy rainfall with the possibility of causing flooding are possible into the Ohio Valley and Mid-Atlantic region as well. On Saturday, heavier rainfall totals are expected across the Ohio Valley, with a risk of flash flooding there. \EsentR{1, R2, R3}{The severe weather threat is forecast farther south across the Southeast and Southern Appalachians.}

The front will divide warmer than average temperatures to the east of it and cooler than average temperatures to the west. The Southeast and Tennessee Valley will be 5 to 10 degrees above average and the Mid-Atlantic region 10 to 15 degrees above average for high temperatures on Friday. A backdoor front across the northern Mid-Atlantic on Friday will keep the Northeast cooler than average. Behind the main front, temperatures are forecast to be 5 to 15 degrees cooler than average in the Southern/Central Plains and into the Middle Mississippi Valley. As the front moves eastward on Saturday, the Ohio Valley could see high temperatures 5 to 10 degrees below average.

\EsentBg{3}{Farther west, scattered thunderstorms are forecast to develop on Saturday in the Southern and Central High Plains, which could be severe.} \EsentBg{4}{A cold front coming into the Northern Plains late Friday will progress into the Upper Midwest on Saturday, causing showers along it.} \EsentBg{2}{Temperatures are expected to cool below average in the Northern Rockies and Plains with this front, which could lead to light snow in those areas on Saturday.} The rest of the West will warm and have above average temperatures for the most part by Saturday, and generally remain dry ahead of a low approaching California by Sunday morning.
\end{tcolorbox}
\caption{Report from Figure~\ref{fig:reportviz} with every \mistraljudge claim labeled SUPPORTED (\textcolor{ClaimSup}{\textbf{S}}), REFUTED (\textcolor{ClaimRef}{\textbf{R}}), or NEUTRAL (\textbf{N}). SUPPORTED and REFUTED claims are shaded green and red, and the reference sentences that support or refute them carry the same tag and shading.}
\label{fig:judgeviz-mistral}
\endgroup
\end{figure}

Together with the \oss labels in Figure~\ref{fig:reportviz}, Figures~\ref{fig:judgeviz-gemma} and~\ref{fig:judgeviz-mistral} show that all three judges support the snow-shower and southern-Plains storm claims, which have direct counterparts in the reference, but differ in how much evidence they require elsewhere. \gemma additionally supports all four claims that \oss leaves NEUTRAL by accepting partial or indirect evidence. For example, it supports the northern-Plains trough and surface low (S1) based on a cold front entering the Northern Plains, and precipitation over the Pacific Northwest and northern Rockies (S7) based on light snow in the Northern Rockies, although the reference keeps most of the West dry. \mistraljudge splits compound sentences into 16 narrower claims and supports only four of them. All three judges refute the sentence describing the Southeast and much of the South as largely dry. Claim granularity and support thresholds therefore change both the claim counts and their labels, so absolute Discussion F1 reflects judge behavior as well as report quality.

\section{Prompts}
\label{app:prompts}

\subsection{Report-stage prompt}
\label{app:agent-prompt}
The report-stage prompt contains a system message and a user message; the cutoff and rejection messages follow. The abridged version retains the action format, the grounding requirement (the prompt's tool-evidence contract), and report style. Brackets mark omitted or summarized content, including the user message that carries the task request. \Direct and \Curr use the same prompt for training and evaluation.

\begin{promptbox}{{Report-stage prompt (abridged, with omissions in brackets)}}
\scriptsize\raggedright\setlength{\parindent}{0pt}\setlength{\parskip}{0.3\baselineskip}
{\rmfamily\itshape [system message]}\newline
You are a weather expert agent.\newline
At each turn, respond with exactly one block: either Python code in \textless{}execute\textgreater{} tags or the final answer in \textless{}solution\textgreater{} tags. Use additional execution turns when they can refine the answer.\newline
{\rmfamily\itshape [tool documentation, data layout, units, indexing rules, plotting guidance, and the execution signature]}

TOOL-EVIDENCE CONTRACT for this task:\newline
- Before a \textless{}solution\textgreater{} is accepted, successfully access the dataset and obtain forecast fields with the forecaster. A premature solution is rejected and costs a turn.\newline
- Every response must contain exactly one tag, never both.\newline
- Base the final report on observations from the current episode.\newline
- After evidence-bearing calls, the environment appends a server-computed [verified evidence] summary. Optional tools should be used only when they materially change the analysis.

FINAL REPORT STYLE -- applies only inside \textless{}solution\textgreater{}:\newline
Write approximately 250-350 words in 3-5 continuous prose paragraphs.\newline
Do not use titles, headings, lists, or separator lines. Begin with the dominant forecast pattern, organise the report around regional and temporal evolution, connect atmospheric mechanisms to impacts, and state uncertainty where evidence is limited. Use numbers selectively and do not describe the tool-use process.\newline
{\rmfamily\itshape [user message: atmospheric variables, time window, calendar context, interaction rules, tool documentation, data schema, units, execution rules, and the complete CONUS weather-report request]}\newline
{\rmfamily\itshape [cutoff policy: at most 12 working turns followed by one forced final-answer turn]}
\end{promptbox}

The environment uses the following messages at cutoff or after an invalid response. The last line is the no-report sentinel, which replaces a missing, too-short, or ungrounded report before offline judging.

\begin{promptbox}{{Forced final turn and rejection messages (opening sentences)}}
\footnotesize\raggedright\setlength{\parindent}{0pt}\setlength{\parskip}{0.3\baselineskip}
{\rmfamily\itshape [grounding requirement met]} Hard cutoff reached. Do not write more code. Your next response must be exactly one \textless{}solution\textgreater{} block containing your complete forecast discussion, closed with \textless{}/solution\textgreater{}.

{\rmfamily\itshape [grounding requirement not met]} Hard cutoff reached. Do not write more code. Required tool evidence was not completed this episode (successful dataset access and a successful forecast were both required), so a forecast discussion cannot be scored. {\rmfamily\itshape [followed by the request for one final \textless{}solution\textgreater{} block]}

{\rmfamily\itshape [premature solution]} Your \textless{}solution\textgreater{} was not accepted yet: the final report must be based on evidence gathered in this episode.

{\rmfamily\itshape [neither tag]} No \textless{}execute\textgreater{} block and no \textless{}solution\textgreater{} block found.

{\rmfamily\itshape [no-report sentinel]} Unable to provide a more detailed answer from the available analysis.
\end{promptbox}

\subsection{Rubric judge prompt}
\label{app:rubric-prompt}
At temperature zero, the rubric judge receives the system line and user prompt below. The inserted rubric contains the complete criteria and scoring anchors from Appendix~\ref{app:rubric}.

\begin{promptbox}{Rubric judge prompt}
\footnotesize\raggedright\setlength{\parindent}{0pt}\setlength{\parskip}{0.3\baselineskip}
{\rmfamily\itshape [system]} You are a meteorology evaluation expert. You score forecast discussions strictly against a rubric. Be terse, precise, and consistent.

{\rmfamily\itshape [user]} Score the following answer against the rubric.\newline
Use the integer 1-10 scale defined by the anchors (1 = worst, 10 = best). Be strict and consistent.\newline
For EACH criterion, give the integer score AND a 1-2 sentence explanation that\newline
\mbox{}~~- cites concrete evidence from the answer (or its absence), and\newline
\mbox{}~~- explains why that evidence places the answer at that anchor.

RUBRIC DESCRIPTION (what this rubric measures): \{RUBRIC DESCRIPTION\}\newline
RUBRIC: \{RUBRIC JSON\}\newline
QUESTION: \{QUESTION\}\newline
REFERENCE (ground truth meteorologist discussion): \{REFERENCE\}\newline
ANSWER TO SCORE: \{ANSWER\}

{\rmfamily\itshape [output format: JSON with an integer score and a brief explanation per criterion, plus one overall sentence; the keys must match the rubric criterion names]}
\end{promptbox}

\subsection{Claim extraction and claim evaluation prompts}
\label{app:entailment-prompts}
Discussion F1 uses one prompt to extract claims and another to label each claim against the other text (the reference for report claims and the report for reference claims). Both prompts are reproduced verbatim as format templates: single braces mark input slots, and doubled braces render as single literal braces.

\begin{promptbox}{Claim extraction}
\footnotesize\raggedright\setlength{\parindent}{0pt}\setlength{\parskip}{0.3\baselineskip}
You are given a meteorological discussion. Please extract all the main claims and present it in compact bullet points.

Discussion: \{\}

Respond with a json list of strings as shown below:\newline
\{\{\newline
\mbox{}~~~~"points": [\newline
\mbox{}~~~~~~~~"point\_1",\newline
\mbox{}~~~~~~~~"point\_2",\newline
\mbox{}~~~~~~~~...\newline
\mbox{}~~~~]\newline
\}\}
\end{promptbox}

\begin{promptbox}{Claim evaluation}
\footnotesize\raggedright\setlength{\parindent}{0pt}\setlength{\parskip}{0.3\baselineskip}
You are a climate expert that annotates if a given claim is either SUPPORTED, REFUTED or NEUTRAL with respect to the presented report. The evidence is in the form of a meteorological discussion.

Use the following definitions:

\textcolor{ClaimSup}{{\fontfamily{lmtt}\fontseries{b}\selectfont SUPPORTED:}} There is at least one statement in the report that directionally agrees with the claim, i.e. the claim could roughly be used as a substitute for some statement(s) in the report.

{\fontfamily{lmtt}\fontseries{b}\selectfont NEUTRAL:} There is not material information in the report related to the claim.

\textcolor{ClaimRef}{{\fontfamily{lmtt}\fontseries{b}\selectfont REFUTED:}} There is at least one statement in the report that directionally disagrees with the claim, i.e. the report and the claim can not both hold true simultaneously.

You will be provided the following as the input:

Claim: \{claim\}

Evidence: \{report\}

Respond with just one word: SUPPORTED, NEUTRAL, or REFUTED.
\end{promptbox}

\end{document}